\documentclass[10pt,twocolumn,letterpaper]{article}

\usepackage[pagenumbers]{cvpr} %
\usepackage{comment}

\usepackage{graphicx}
\definecolor{cvprblue}{rgb}{0.21,0.49,0.74}
\usepackage[pagebackref,breaklinks,colorlinks,allcolors=cvprblue]{hyperref}

\def\paperID{*****} %
\def\confName{CVPR}
\def\confYear{2026}

\title{What Happens to Accuracy When Photo Lineups Contain Non-Mated Rank-One Images From Large Galleries?}

\author{Genesis Argueta\\
University of Notre Dame\\
Notre Dame, USA\\
{\tt\small gargueta@alumni.nd.edu}
\and
Kevin W. Bowyer\\
University of Notre Dame\\
Notre Dame, USA\\
{\tt\small kwb@nd.edu}
\and
Michael King\\
Florida Institute of Technology\\
Melbourne, USA\\
{\tt\small michaelking@fit.edu}
\and 
Jayeeta Dhar\\
University of Notre Dame\\
Notre Dame, USA\\
{\tt\small jdhar@nd.edu}
}
\begin{document}
\maketitle
\begin{abstract}
One-to-many facial identification is commonly used to match a probe image from surveillance video against a gallery of driver’s license and/or booking photos.
The algorithm’s rank-one image from the gallery, or a human examiner’s selection from the algorithm’s top-ranked images, may then be placed in a photo lineup shown to a witness.
Witness selection of the gallery image in the photo lineup may then lead directly to the person in the gallery image being arrested.
This facial identification process is involved in a number of wrongful arrests.
This work specifically examines whether the probability of a witness making an incorrect identification increases with the size of the gallery searched. 
We compare photo lineup accuracy when the ``suspect’’ image is drawn from galleries of 500, 5,000, and 24,000 images. 
We find that larger galleries increase both the likelihood of a witness making an incorrect identification and their confidence in that (incorrect) identification.
These results raise questions of whether an image resulting from such a facial identification process should be used in photo lineups and of whether results of a photo lineup alone should constitute probable cause for arrest.

\end{abstract}
    
\section{Introduction}
\label{sec:intro}
The use of facial recognition technology in law enforcement has expanded rapidly.
Automated one-to-many facial identification systems now routinely search across increasingly large databases---from tens of millions of images \cite{aclu_williams} to over 70 billion images \cite{clearview2026overview}---to generate investigative leads. However, a critical question remains largely unexamined: does the size of the database searched affect the fairness of subsequent photo lineups shown to witnesses?

This paper focuses on a common scenario. A probe image, often obtained from surveillance video, is used in one-to-many facial identification. The result of the facial identification is a candidate suspect image and identity, typically referred to as an ``investigative lead’’.   The image from the one-to-many facial identification, or possibly another image of that person, is incorporated into a photo lineup alongside ``filler'' (known innocent person) images, and the lineup is presented to a witness.  In many of the known instances of wrongful arrest, a witness selecting the suspect image from the lineup appears to be the only `evidence’ leading to the `suspect’ being arrested.  

If larger galleries systematically result in suspect images that are stronget look-alikes for the probe image than are the filler images, even when the person in the probe image is not in the gallery, then the resulting photo lineup increases the risk of innocent persons being incorrectly identified from the lineup.  Current photo lineup construction procedures vary substantially across jurisdictions \cite{NRC2014}.  To our knowledge, no jurisdiction conditions how the photo lineup is conducted, or how its result is interpreted, based on the size of the gallery searched to obtain the suspect image.  However, the state of Indiana has recently passed a new law that says, ``if facial recognition technology is used to identify a suspect, a law enforcement agency …  may not conduct a lineup unless there is other evidence, in addition to the use of facial recognition technology, to support a belief that the suspect committed the crime under investigation’’ \cite{Indiana2025}.  Had a law of this type been in effect already, many of the known instances of wrongful arrest associated with one-to-many facial identification might have been prevented.

The importance of understanding the potential bias in using one-to-many facial identification to generate a suspect image for a photo lineup is underscored by the wrongful arrests in which a witness made an incorrect identification from such a lineup. Examples include the wrongful arrests of 
Beau Burgess \cite{Burgess_article}, 
Robert Dillon \cite{Dillon_article}, 
Chris Gatlin \cite{Gatlin_article},
Michael Oliver \cite{Oliver_article}, 
Jalil Richardson \cite{Richardson_article}, 
Alonzo Sawyer \cite{Sawyer_article}, 
Robert Williams \cite{R_Williams_article},
Trevis Williams \cite{T_Williams_article}, 
and
Porcha Woodruff \cite{Woodruff_article}.
These cases illustrate how errors introduced during facial identification may propagate into subsequent eyewitness identification procedures and ultimately contribute to wrongful arrests.
\begin{figure}[t]
    \centering
    \includegraphics[width=0.56\columnwidth]{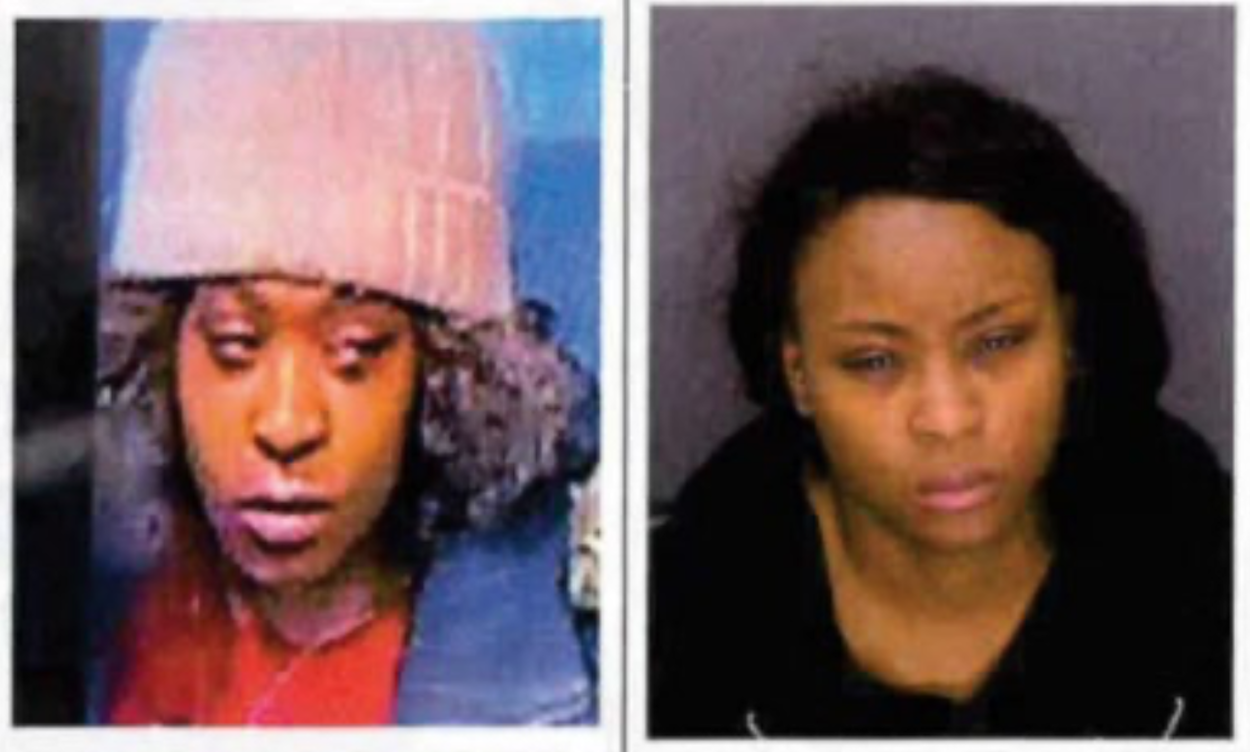}
    ~~
     \includegraphics[width=0.25\columnwidth]{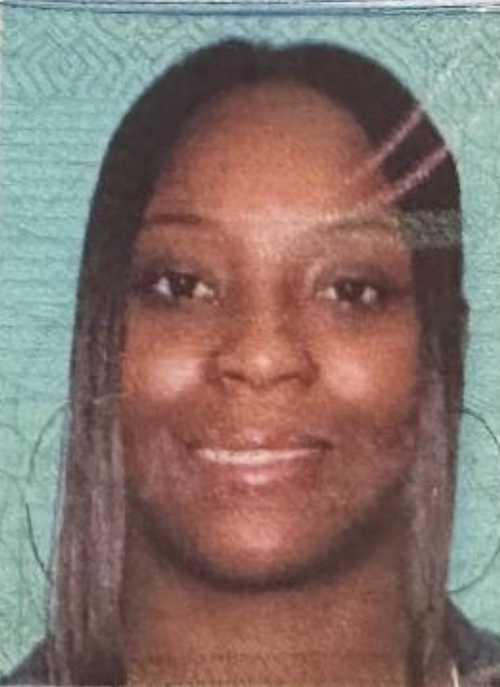}
\\
(a) left: probe + gallery image; right: recent DL image
\\[6pt]
\includegraphics[width=0.85\columnwidth]{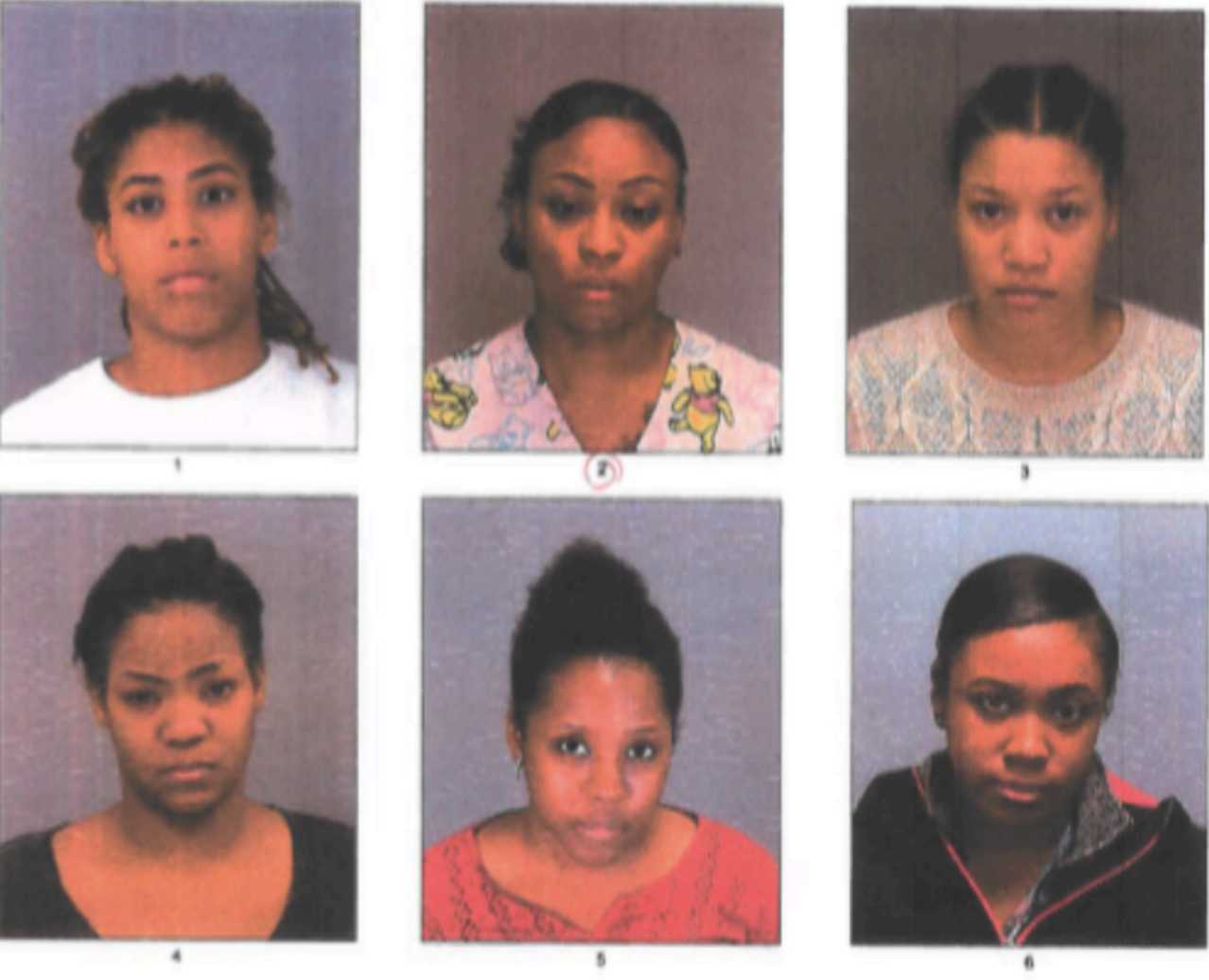}
\\
(b) lineup show to eyewitness, with older DL image
\caption{
Photo lineup used in the Porcha Woodruff wrongful arrest \cite{Woodruff_court}.  Facial recognition using a probe from surveillance video returned an old booking photo, which was not used in the lineup.  A newer driver's license photo existed, but was not used. An older driver's license photo (second image in top row of lineup) was used in the lineup, and was selected by the victim.
Woodruff was eight months pregnant at the time of the crime, something not mentioned in the victim's description of the culprit; charges were subsequently dismissed. This case illustrates the investigative workflow studied in this paper, and the complexity of actual practice from surveillance video to witness viewing of a photo lineup.
}
    \label{fig:porcha}
\end{figure}

This paper makes four contributions to the understanding of facial identification and eyewitness accuracy. First, we provide the first empirical investigation of how gallery size in automated one-to-many facial identification affects the fairness and accuracy of subsequent photo lineups shown to eyewitnesses. Second, we demonstrate that suspect images retrieved from larger galleries are systematically more similar to the probe image than standard fillers, creating lineups that are structurally biased toward a witness incorrectly selecting an innocent suspect. Third, we show that this increased similarity not only leads to the witness being more likely to incorrectly select an innocent person, but also to the witness being more confident about their incorrect decision. Finally, we discuss the implications of our results for current lineup construction policy.
\begin{figure}[t]
\centering

\textbf{(a) Suspect-present lineup}

\vspace{2mm}

\includegraphics[width=0.25\columnwidth]{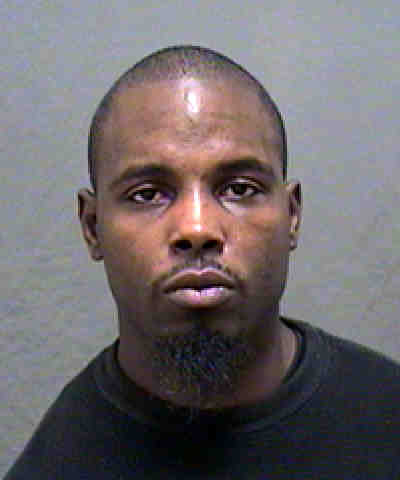}
\includegraphics[width=0.25\columnwidth]{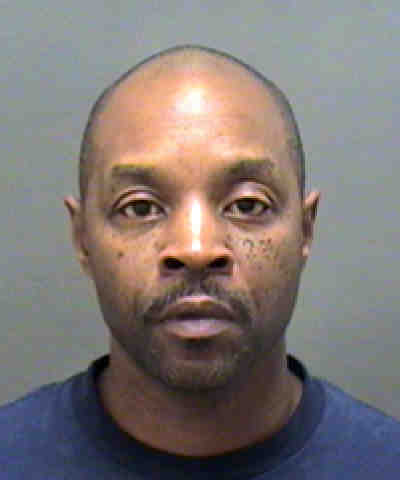}
\includegraphics[width=0.25\columnwidth]{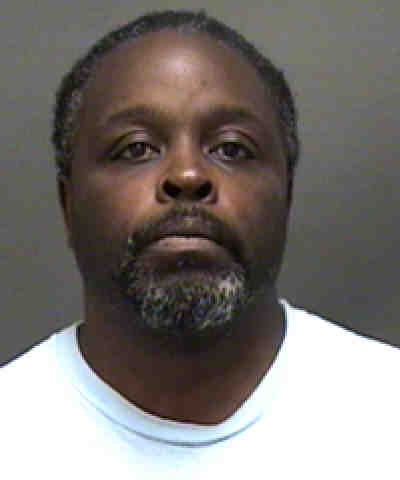}

\includegraphics[width=0.25\columnwidth]{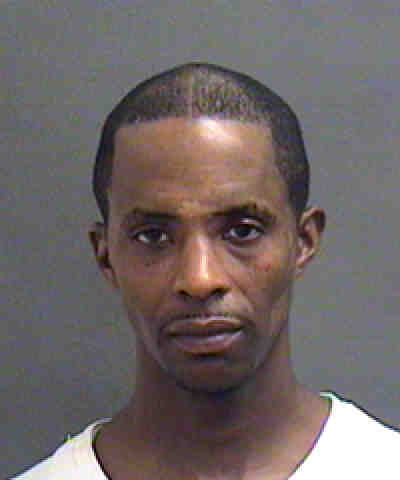}
\includegraphics[width=0.25\columnwidth]{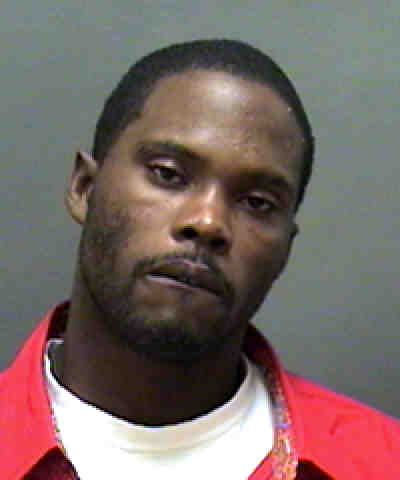}
\includegraphics[width=0.25\columnwidth]{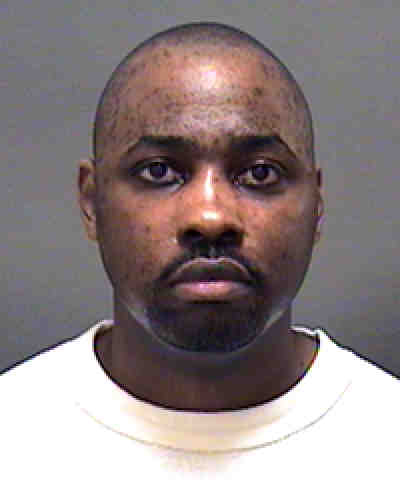}

\vspace{3mm}

\textbf{(b) Rank-one non-mate lineup}

\vspace{2mm}

\includegraphics[width=0.25\columnwidth]{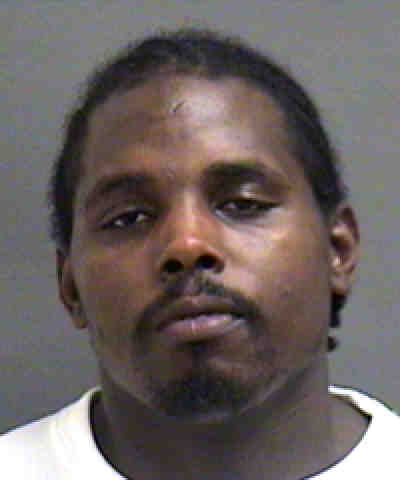}
\includegraphics[width=0.25\columnwidth]{figs/experiment1/lineup4/068291_03M50.JPG}
\includegraphics[width=0.25\columnwidth]{figs/experiment1/lineup4/114513_06M48.JPG}

\includegraphics[width=0.25\columnwidth]{figs/experiment1/lineup4/161494_08M36.JPG}
\includegraphics[width=0.25\columnwidth]{figs/experiment1/lineup4/231716_07M27.JPG}
\includegraphics[width=0.25\columnwidth]{figs/experiment1/lineup4/259325_01M35.JPG}

\caption{
Example lineup construction. Panel (a) contains the true suspect image and five fillers. Panel (b) replaces the suspect with the highest-ranked non-mated image retrieved from a gallery of approximately 24,000 images. This work investigates whether highly similar non-mated images increase the likelihood of erroneous identifications in photographic lineups.
}
\label{fig:lineup_example}
\end{figure}

\section{Literature Review}
\label{sec:litreview}
Prior work relevant to this study falls into three broad areas: the effect of gallery size on automated face recognition performance, the accuracy of human decision-making when evaluating algorithm-generated candidate lists, and the accuracy of eyewitness identifications from photo lineups.

\subsection{Gallery Size and Automated Face Recognition}
A consistent finding across the literature is that identification accuracy degrades as gallery size increases, because larger galleries make it more likely that highly similar non-matching individuals appear among the top-ranked candidates. Johnson et al.\ \cite{johnson2003gallery} demonstrated that rank-one match quality declines with gallery size and developed methods for estimating large-gallery performance from smaller galleries, though their work focused solely on algorithmic metrics. Friedman et al.\ \cite{friedman2022gallery} similarly found that identification becomes progressively harder as the number of distractor identities grows. Vareto et al.\ \cite{vareto2016largegalleries} provided a systematic quantification of this effect across galleries ranging from 1{,}000 to 64{,}000 subjects, showing that rank one performance decreases approximately linearly with the logarithm of gallery size. Pangelinan et al.\ \cite{pangelinan2024analyzing} directly demonstrated the underlying mechanism: as the number of enrolled identities increases, the non-mated similarity score distribution shifts toward higher values while the mated distribution remains largely unchanged. As they note, the larger the number of persons in the gallery, the more likely a strong ``look-alike'' will be found.\footnote{More specifically, it is the number of persons in the same demographic group as the person in the probe image that most directly affects accuracy.  For example, doubling the number of males in the gallery will reduce accuracy for probe images of males, but not make a noticeably change in accuracy for probe images of females.}
Taken together, these studies establish that gallery size directly governs the similarity of the highest-ranked non-mate images returned by a face recognition system.

\subsection{Examiner accuracy when evaluating algorithm-selected candidate lists}
A separate body of work examines how well human examiners can identify the correct match when presented with a candidate list generated by an automated system. White et al.\ \cite{Whiteetal2015} found that professional facial examiners achieved roughly 70\% correct detection when comparing a probe image to the eight most-similar passport images returned by a pre-deep-learning face matcher from a gallery exceeding one million images. Clothier et al.\ \cite{Clothier2024} extended this line of inquiry by comparing older and newer (deep-learning-based) matchers, finding that more accurate algorithms paradoxically make the examiner's task harder: the higher-quality candidate lists they produce contain more confusable non-mates, reducing both correct selections when the target is present and correct rejections when it is absent. 
Heyer \etal \cite{Heyer2018} examine how the accuracy of examiners is affected by the size of the candidate list.  In one their experiments, the examiners were Australian government employees who worked as face examiners, and accuracy was compared for evaluating candidate lists of  10, 50 and 100 images.  
(For context, the candidate list considered by the Michigan State Police in the Robert Williams wrongful arrest case was 243 images, and the examiner selected the number 9 image in the algorithm’s ranking \cite{aclu_williams}.)
They found that larger candidate lists increased false alarms and decreased hit rate.  Interestingly, they also found that faster judgments were more accurate.  
These studies deal with examiner accuracy in selecting a suspect image from an algorithm-generated candidate list, whereas our work focuses on witness accuracy in identifying the suspect image from a photo lineup.

\subsection{Eyewitness identification from photo lineups}

The reliability of eyewitness identification has been a central concern in both psychological research and legal policy \cite{wells1995, NRC2014, Wells2020b}. Eyewitness memory is known to be malleable and susceptible to post-event influences \cite{loftus2005}, and identification accuracy varies substantially across procedural conditions and demographic factors \cite{cutler1987, MeissnerBrigham2001}. Fitzgerald et al.\ \cite{Fitzgerald2021} documented considerable variation in eyewitness identification procedures and outcomes across different countries and legal systems. Kelso et al.\ \cite{Kelsoetal2025} demonstrated that an AI tool analyzing the content of witness statements after a lineup decision can predict identification reliability more accurately than human evaluators alone, particularly when witnesses provide feature-based or recognition-based justifications.

Bergold and Heaton \cite{Bergold2018} examine the issue of database size in the selection of filler images to go into a lineup for a given a suspect image.  Their work gets at the issue of filler images of known innocents becoming too similar to the suspect, making the identification task more difficult for the eyewitness, causing the eyewitness to incorrectly select filler images more often.  Our work focuses on an earlier step and a different question.  We focus on the issue of larger gallery size resulting in selection of a suspect image that is more similar to the actual culprit image.  The issue we address is that larger gallery size results in the suspect being so similar to the culprit that it may cause the eyewitness to incorrectly select the suspect when in fact they are not the culprit.  

Eyewitness identification studies also vary considerably in how they induce memory of a suspect prior to the lineup task. The most common approach involves showing participants a mock crime video in which the perpetrator's face is visible for a controlled duration to simulate the incidental encoding that occurs when witnessing a real crime \cite{fitzgerald2025eli}. Other studies use briefer static exposures, presenting a single photograph of the target face for as little as 5 seconds before the identification test \cite{meyer2023interactive}. 
These studies illuminate factors that influence lineup accuracy, but none connects lineup construction to the one-to-many facial identification by which a suspect image was retrieved from a large gallery.

Wells \etal \cite{Wells2020} discuss the most recent American Psychological Association guidelines for eyewitness identification procedures.  Of the nine recommendations, one that is particularly relevant to our work is the Evidence-Based Suspicion Recommendation – “There should be evidence-based grounds to suspect that an individual is guilty of the specific crime being investigated before including that individual in an identification procedure …” \cite{Wells2020}.  In the context of our work, the question is whether an individual being selected by one-to-many facial identification as a possible match to a probe image can, by itself, be considered as evidence-based grounds to suspect that the individual is guilty.
A law that took effect in the state of Indiana in 2026 addresses this specific point \cite{Indiana2025} - ``If facial recognition technology is used to identify a suspect, a law enforcement agency ... may not conduct a lineup unless there is other evidence, in addition to the use of facial recognition technology, to support a belief that the suspect committed the crime under investigation.''
Unfortunately, there is not yet any federal law addressing this point.

\subsection{Summary and gap addressed by this work}
There is substantial literature on the accuracy of automated one-to-many facial identification \cite{FRVT_Identification, johnson2003gallery, vareto2016largegalleries, friedman2022gallery, pangelinan2024analyzing}, on the accuracy of decisions made by face examiners evaluating an algorithm's candidate list \cite{Whiteetal2015, Clothier2024}, and on the accuracy of identifications made by witnesses presented with a photo lineup \cite{wells1995, NRC2014, Kelsoetal2025}. There is little if any literature specifically on the accuracy of photo lineup identifications when the suspect image in the lineup is itself the product of one-to-many facial identification from a large gallery. Critically, the gallery-size research reviewed above predicts that larger galleries will produce more similar non-mated images, yet no prior work examines whether that increased non-mate similarity then compromises the fairness or accuracy of a subsequent eyewitness lineup. This is the gap that the present study addresses.

\section{Experimental Design}

\subsection{Dataset and Subject Selection}

We used subsets of the MORPH dataset \cite{morph, uncw_morph} to construct the experimental photo lineups. MORPH is a large-scale collection of mugshot-style images gathered from law enforcement records, spanning many thousands of unique subjects and containing multiple images per subject collected over time. Each image is accompanied by demographic metadata (e.g., age, sex, and race), and the images themselves reflect the standardized, frontal, controlled-pose format typical of booking photographs. Because the images are mugshots rather than casual or in-the-wild photographs, MORPH is uniquely suitable for studying face identification under forensic-like conditions, such as photo lineup construction, where consistency of pose and capture conditions across images is important for isolating the effects of interest.

Two curated releases of MORPH were used to build the galleries described in Section 3.2. The 500-image and 5,000-image galleries were drawn from a curated version of MORPH v3 \cite{Albiero_2022}. MORPH v3 did not contain enough eligible images to support the largest (~24,000-image) gallery, and so this gallery was drawn from a later and larger version of MORPH (v5) that was cleaned similarly to MORPH v3.

From the full dataset, we restricted our analysis to the adult African-American male cohort. This subset was selected for two reasons. First, it contains the largest number of eligible subjects and images of any demographic group in MORPH, which was necessary for constructing galleries of increasing size (500, 5,000, and approximately 24,000 images; see Section 3.2) from a sufficiently large pool within a single demographic group, so that gallery size, rather than demographic composition, was the variable being manipulated. Second, identifying suspect identities with enough usable images per person (see next paragraph) required a cohort large enough that this additional filtering would still leave an adequate number of candidate identities.

All images within this cohort were first restricted to those subjects at least 18 years of age; subjects for whom every available image fell below this age threshold were excluded from consideration entirely. Then, we filtered out all subjects with fewer than six available images. This threshold ensured that each eligible suspect identity had enough images to support every experimental component that draws on a single identity: one probe image for face matching, four images for the memory inducing phase, and at least one additional image reserved for the target-present lineup condition (see next paragraph). Subjects with fewer than six images could not simultaneously supply all of these components and were therefore excluded.

From the remaining eligible subjects, we randomly selected 8 identities to serve as suspect identities across experiments. Eight identities were selected to support a target of six suspect identities' worth of lineups per participant, a number chosen to keep the overall experimental session under one hour. The remaining two identities were selected as a buffer, to be used as replacements in case any of the initial six lineups were found to be problematic (e.g., due to an issue with a filler or probe image identified during piloting or review). For each selected suspect identity, one image was designated as the probe image, four images were used in the memory inducing phase, and the remaining image(s) were reserved for constructing the target-present lineup condition.

\subsection{Gallery Construction}

To construct non-suspect candidate pools, we created three independent image-level galleries by uniformly sampling images (not identities) from the full pool of eligible images, prior to selecting any suspect identities. Because sampling was performed at the image level, multiple images from the same individual could appear within a single gallery, and because the 500-image and 5,000-image galleries were each sampled independently from the same underlying pool, the smaller gallery is not guaranteed to be a subset of the larger one. An ArcFace embedding (ResNet-100 backbone; \cite{deng2019arcface}) was computed for every image in each gallery in advance of suspect selection, so that similarity search against a suspect's probe image (Section 3.3) could later be performed efficiently.

We constructed:
\begin{itemize}
    \item a 500-image gallery,
    \item a 5,000-image gallery,
    \item and a full gallery consisting of approximately 24,000 images (the entire eligible MORPHv5 cohort after filtering).
\end{itemize}

Suspect identities were then selected one at a time, across eight experiments. In each experiment, a suspect identity was randomly chosen from the pool of subjects that (a) had not already served as a suspect in a previous experiment, and (b) had at least six eligible images. Similarly, once a subject's image had been used as a filler (Section 3.4) or as a suspect in a given experiment, that subject was excluded from serving as a suspect or filler in any subsequent experiment; no subject identity was reused across the eight experiments in either role.

Each suspect identity resulted in the selection of 18 images used across all experimental conditions. These consisted of: (1) one suspect probe image used for face matching, (2) four probe images used in the lineup tasks (one true target-present suspect image and three rank-one non-mates, one from each gallery), (3) four additional images of the suspect used in the matching exercise, (4) four non-suspect images used in the matching exercise, and (5) five filler images that remained consistent across lineup conditions.

\subsection{Face Recognition Search and Rank-One Non-Mate Selection}

For each suspect identity, a probe embedding was computed from the designated probe image using a pretrained face recognition model, ArcFace with a ResNet-100 backbone \cite{deng2019arcface}. This embedding was compared, using cosine similarity, against the precomputed embeddings of every image in each gallery, excluding any images belonging to the suspect's own identity or to a subject already used as a suspect or filler in a prior experiment.

For each gallery, we selected the \textit{rank-one non-mate image}, defined as the image with the highest cosine similarity to the probe image that does not belong to the suspect identity (and, per the exclusion criteria above, does not belong to any subject already used elsewhere in the study). This procedure was performed separately for the 500-image and 5,000-image galleries (and, for the full ~24,000-image gallery, yielding three distinct non-mates per suspect identity. The purpose of the three galleries is to examine how the accuracy of six-pack lineup decisions varies with increasing gallery size.

\subsection{Photo Lineup Construction}

A photo lineup is a "six pack" of images. Six images in a photo lineup is common in the United States, although other countries may use larger sizes \cite{Fitzgerald2021}. Each lineup consisted of one probe image and five filler images. The same set of filler images was used across all lineup conditions for a given suspect identity to ensure comparability across conditions; a different set of filler images was selected for each suspect identity to ensure appropriate fillers for that identity.

Filler images (for both the training phase and the lineup conditions) were selected from the 5,000-image gallery, ranked by cosine similarity to the suspect's probe embedding, after excluding the suspect's own images and any subject already used as a suspect or filler in a prior experiment. To avoid trivial exclusion of a filler image based on obvious artifacts, candidate filler images were screened to remove images with conspicuous scars, markings, or other highly distinguishing features when necessary. Fillers were selected from mid-ranked similarity positions, excluding the highest-ranked candidates, to maintain moderate difficulty and variability across lineups: the five lineup filler images were the 21st through 25th most similar remaining candidates in the 5,000-image gallery. The filler images in the memory inducing task were also selected from the 5,000-image gallery where they were the 26th-29th most similar remaining candidate.

\subsection{Experimental Conditions}

For each suspect identity, four lineup conditions were constructed:

\begin{itemize}
    \item \textbf{Lineup A (target-present):} includes the true suspect image plus five fillers.
    \item \textbf{Lineup B (500-gallery condition):} includes the rank-one non-mate image from the 500-image gallery plus five fillers.
    \item \textbf{Lineup C (5,000-gallery condition):} includes the rank-one non-mate image from the 5,000-image gallery plus five fillers.
    \item \textbf{Lineup D (24,000-gallery condition):} includes the rank-one non-mate image from the full ~24,000-image gallery plus five fillers.
\end{itemize}

This design isolates the effect of gallery size on the quality of the highest-similarity non-mate, while holding constant the probe image, filler construction procedure, and overall lineup structure across conditions.
\section{Experimental Procedure}

\subsection{Participant Tasks Structure}

Each participant completed six lineup identification tasks corresponding to six different suspect identities. The order in which suspects were presented was randomized for each participant to control for order effects.

Prior to viewing each lineup, participants completed a memory inducing phase designed to establish visual familiarity with the suspect. We refer to this encoding phase as the "memory inducing phase" throughout this paper. Participants were given one image of the suspect identity to hold and examine. They were then presented with eight images arranged randomly on a table and asked to sort them into two piles: images depicting the suspect and images depicting other individuals. Four of the eight images were of the suspect identity, and the other four were of different individuals. The four other images were the 26th, 27th, 28th, and 29th most similar images to the probe image. Participants could not move onto the lineups until they successfully made the two piles. This process typically took between one to five minutes. This matching task simulated the cognitive process of encoding a suspect's appearance into memory and establishing familiarity with their facial features, analogous to a witness observing a perpetrator during a crime.

Following the memory-induction phase, participants were given a one-minute break before proceeding to the lineup identification task. During this interval for their first lineup, participants were read the photo lineup instructions aloud; for subsequent lineups, they were permitted to use their phones, drink water, or use the restroom. These instructions were adapted from the official protocol used by the Multnomah County District Attorney's Office and the Portland Police Bureau~\cite{mcda2015eyewitness}.

During the lineup task, participants viewed the six lineup images sequentially—one at a time—and were asked to indicate whether each image matched their memory of the suspect from the training phase. The order of the presentation of the six images were also randomized across each lineup. Participants could not revisit images once a decision was made. 

For each image viewed, participants made a binary yes/no decision regarding whether the image depicted the suspect. Regardless of their responses, all participants viewed all six images in the lineup before the procedure concluded. If a participant responded “yes” to any image, the experimenter immediately asked them to rate their confidence on a scale from 1 to 10, with 1 being the least confident and 10 being the most confident, and explain the reasoning behind their selection. If a participant responded “no” to all six images in the lineup, indicating that the suspect was not present, the experimenter asked them to rate their confidence that the suspect was indeed absent from the lineup. Participants were permitted to change their responses as they progressed through the lineup; however, they were not allowed to revisit previously viewed images. For example, a participant who initially identified the first image as the suspect could later revise their response after viewing subsequent images. In such cases, the experimenter overwrote the participant’s previous response and repeated the confidence and reasoning questions for the newly selected image

The sequential six-image lineup procedure used in this study reflects standard practice employed by law enforcement agencies across the United States \cite{wisconsin_eyewitness_2026}. It is worth noting, however, that lineup procedures are not uniform across jurisdictions. Some agencies present all lineup images simultaneously rather than sequentially, some use more than six images, and some permit witnesses to make multiple passes through the lineup before rendering a final decision \cite{NRC2014}. The present study does not evaluate all possible procedural variants; rather, it adopts a common and well-documented procedure as a controlled baseline from which to examine the effect of gallery size on eyewitness identification accuracy.

\subsection{Experimental Phases and Condition Balancing}

The experiment was conducted in two phases to ensure comprehensive data collection across all lineup conditions while managing participant workload.

In Phase 1, participants viewed lineups for Suspects 1 through 6 (images at the end of paper). Each participant encountered two instances each of Lineup A (target-present), Lineup B (500-gallery rank-one non-mate ), and Lineup C (5000-gallery rank-one non-mate). The assignment of suspects to conditions was counterbalanced across participants using a rotation scheme (e.g., ABC, BCA, CAB) to ensure that each suspect identity appeared equally often in each experimental condition across the participant pool.

Phase 2 used Suspects 1, 2, 5, 6, 7, and 8, replacing Suspects 3 and 4 from Phase 1. These substitutions were necessary because images from the 24,000-image gallery used in Lineup D had appeared as either filler images or probe images in the lineups for Suspects 3 and 4 during Phase 1. To avoid any potential recognition or contamination effects, these two suspects were replaced with fresh suspect identities (7 and 8) that had no prior exposure to images from the large gallery. Participants in Phase 2 viewed two instances each of Lineup B, Lineup C, and Lineup D, with the same counterbalancing rotation applied.

This two-phase design served multiple purposes. First, it prevented individual participants from experiencing fatigue by limiting each session to six lineups. Second, it ensured that Lineup A (target-present) and Lineup D (24000-gallery) were never directly compared within the same participant, avoiding potential interference effects between conditions where one lineup contains the true suspect and another contains a highly similar non-mate from the largest gallery. Third, the counterbalanced assignment of suspects to lineup conditions prevented confounding effects of specific suspect characteristics with lineup condition.

\subsection{Participant Recruitment and Demographics}

A total of 104 participants were recruited from the University of Notre Dame student participated in both phases. Each experimental session lasted between 40 and 60 minutes, and participants received a \$20 gift card as a participation incentive for their time.

Basic demographic information (gender) was collected but not used in the primary analysis. The study was approved by the University of Notre Dame Institutional Review Board (25-10-9620) , and all participants provided informed consent before beginning the experiment.

\section{Effects of Hairstyles}

During the experiment, we observed that hairstyle consistency appeared to substantially influence participants' ability to recognize suspects across images. In the memory inducing exercises, participants often struggled when images of the same suspect contained significantly different hairstyles. For example, if the initial probe image depicted a suspect with braided hair, participants frequently expressed hesitation or confusion when later viewing images of the same individual with a closely cropped haircut or buzz cut. Observe Figure 10, Figure 16, and Figure 17.
This effect appeared even when other facial features remained relatively stable (Figure 17). Participants sometimes rejected genuine images of the suspect during the sorting task because the hairstyle differed from the hairstyle shown in the original reference image. Informal participant comments suggested that hairstyle was often used as a primary matching cue, particularly during the early stages of the experiment before participants developed more refined comparison strategies.
These observations suggest that easily changeable external features, such as hairstyle, may play a larger role in human face matching than expected. This is especially important in the context of photo lineups, where hairstyle differences between lineup images may unintentionally increase or decrease perceived similarity independent of underlying facial structure.
The influence of hairstyle may also interact with the face recognition system itself. While the ArcFace embeddings are intended to emphasize stable facial characteristics, participants may rely more heavily on superficial visual cues when making rapid identification decisions. 

These findings raise several questions concerning lineup policy and procedure. First, should witnesses viewing a photo lineup be explicitly instructed to focus on stable facial features such as bone structure and facial geometry, rather than changeable attributes like hairstyle? Such instructions could reduce erroneous identifications driven by superficial feature mismatches. Second, our observations about reliance on external cues may compound known challenges in cross-race identification, where witnesses already show reduced sensitivity to within-group facial variation \cite{MeissnerBrigham2001}, raising the question of whether other-race photo lineups should be treated as fundamentally less reliable and subject to additional procedural safeguards. Third, given that participants in our study appeared to improve their comparison strategies over the course of the experiment, it is worth asking whether witnesses should complete a practice lineup prior to the meaningful one, as a form of calibration that could reduce reliance on superficial cues and improve overall accuracy. Future work should more systematically investigate these questions and examine whether controlling hairstyle similarity across lineup images improves identification accuracy.

\section{Anecdotal Observations}

Although the primary focus of this study was quantitative lineup identification accuracy, participants provided qualitative explanations for their decisions and identification strategies. This revealed substantial variation in how participants perceived and compared faces during the task.

A recurring theme in participant comments involved beliefs about cross-race facial recognition ability. Informal discussions with several participants revealed conflicting views on whether the task would be easier if the images viewed belonged to their own racial group. Some participants reported that familiarity with their own racial group improved their ability to distinguish subtle facial differences. For example, a participant from Asia stated that they were better able to differentiate between Asian faces due to greater exposure and experience viewing individuals from that demographic group.

In contrast, other participants expressed the opposite perspective. One participant suggested that individuals from less familiar racial groups could sometimes appear easier to distinguish because the differences between faces seemed more visually significant relative to their own appearance. 

These conflicting perspectives highlight the complexity of the “other-race effect” and suggest that subjective experiences with facial recognition may not always align with existing assumptions about familiarity and identification accuracy.
Participants also described relying on different facial features during identification decisions. Some reported focusing heavily on hairstyles, facial hair, or head shape, while others emphasized more stable facial characteristics such as eye spacing, nose structure, or jawline shape. Several participants noted that their comparison strategies evolved throughout the experiment as they gained experience with the lineup task.

While this experiment was not designed to systematically measure these observations, they provide useful context for interpreting participant behavior and motivate several directions for future work, particularly regarding cross-race identification effects and feature-matching strategies in lineup construction.

\section{Results}
\subsection{Non-mate similarity as a function of gallery size}
To verify that increasing the search gallery produced rank-one non-mates that were progressively more similar to the true suspect — without altering how similar those non-mates were to the fillers — we computed, for each suspect identity and gallery condition, (1) the cosine similarity between the suspect's probe embedding and the rank-one non-mate's embedding, and (2) the mean cosine similarity between that non-mate's embedding and the five filler embeddings.

Suspect-to-non-mate similarity increased with gallery size, from $M = 0.181$ ($SD = 0.084$) in the 500-image gallery to $M = 0.249$ ($SD = 0.098$) in the 5{,}000-image gallery to $M = 0.406$ ($SD = 0.143$) in the 24{,}000-image gallery. A one-way ANOVA confirmed a significant effect of gallery size on this measure, $F(2, 19) = 6.63$, $p = .007$. Pairwise comparisons indicated that the 24{,}000-image condition produced significantly higher suspect-to-non-mate similarity than the 500-image condition ($p = .015$), with a similar but marginally non-significant difference relative to the 5{,}000-image condition ($p = .065$); the difference between the 500- and 5{,}000-image conditions did not reach significance ($p = .183$).

By contrast, non-mate-to-filler similarity remained stable across conditions ($M = 0.138$, $0.123$, and $0.149$, respectively), and a parallel ANOVA revealed no significant effect of gallery size, $F(2, 19) = 0.50$, $p = .615$. Together, these results indicate that larger galleries yielded non-mates that were substantially closer algorithm matches to the suspect, while the non-mate's resemblance to its surrounding fillers --- and thus its relative `fit' within the lineup --- remained unchanged. This validates the central manipulation of the study: differences in identification accuracy across Lineups B, C, and D can be attributed to the increasing similarity between the non-mate and the true suspect, rather than to incidental changes in overall lineup composition.

It is worth noting that the gallery sizes used in this study are substantially 
smaller than those used in operational law enforcement deployments. The images 
in the MORPH dataset are actual police booking photographs, lending realism to 
the gallery, but databases searched in practice can be orders of magnitude 
larger --- a database of 49 million photographs was searched in the wrongful 
arrest of Robert Williams in Michigan \cite{aclu_williams}, and public 
databases in the United States are estimated to hold images of 117 million 
Americans \cite{gao2019face}. If the trend observed here continues 
to such scales, the rank-one non-mate from a facial recognition search 
may effectively be guaranteed to be a doppelganger for the true subject, making the concern raised in this paper considerably more pressing in 
real-world deployments.
\begin{table}[t]
\centering
\caption{Statistical comparisons of suspect-to-non-mate similarity across gallery sizes.}
\label{tab:similarity_pvalues}
\begin{tabular}{lc}
\toprule
Comparison & $p$ \\
\midrule
Omnibus ANOVA, $F(2,19) = 6.63$ & .007$^{**}$ \\
\midrule
500 vs.\ 5,000 & .183~~ \\
500 vs.\ 24,000 & .015$^{*}$ \\
5,000 vs.\ 24,000 & .065~~ \\
\bottomrule
\end{tabular}

\vspace{4pt}
\begin{minipage}{0.8\columnwidth}
\footnotesize
\textit{Note.} Non-mate-to-filler similarity did not differ significantly across gallery sizes, $F(2,19) = 0.50$, $p = .615$; see Figure~3 for means and standard deviations by condition.\\
$^{*}p < .05$. $^{**}p < .01$.
\end{minipage}
\end{table}
\begin{figure}[t]
    \centering
    \includegraphics[width=0.85\columnwidth]{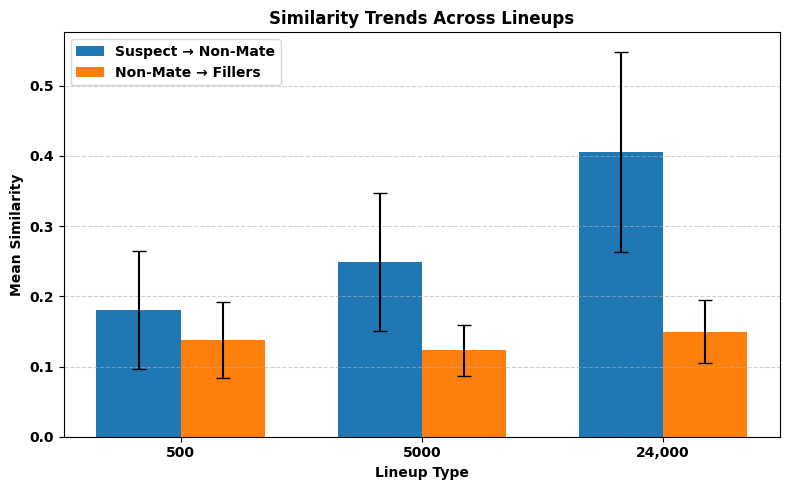}
    \caption{Mean cosine similarity between the suspect probe and
the rank-one non-mate image, and between
the rank-one non-mate and the filler images, for each of the
 three gallery sizes (500, 5,000, and 24,000
images). Error bars represent one standard deviation. Suspect-to-non-mate similarity increased significantly with gallery size ($F(2,19)=6.63$, $p=.007$), while non-mate-to-filler similarity did not differ significantly across conditions ($F(2,19)=0.50$, $p=.615$).}
    \label{fig:similarity_trends}
\end{figure}

\subsection{Overall Accuracy}
Overall identification accuracy differed significantly across the four lineup conditions, $\chi^2$(3, N = 622) = 13.56, p = .004. Accuracy was 53.92\% for Lineup A (target-present), 58.65\% for Lineup B (500-image gallery non-mate), 56.25\% for Lineup C (5,000-image gallery non-mate), and 37.50\% for Lineup D (24,000-image gallery non-mate). This is shown in Figure 4.
Pairwise comparisons indicated that Lineups A, B, and C did not differ significantly from one another (A vs. B: p = .504; A vs. C: p = .790; B vs. C: p = .692). Lineup D, however, was significantly less accurate than each of the other three conditions (D vs. A: $\chi^2$(1) = 4.96, p = .026; D vs. B: $\chi^2$(1) = 11.59, p \textless .001; D vs. C: $\chi^2$(1) = 9.01, p = .003).
For Lineup A, accuracy reflects the proportion of trials on which participants correctly selected the true suspect. For Lineups B, C, and D — each of which substituted the suspect with a rank-one non-mate and therefore contained no true target — accuracy reflects the proportion of trials on which participants correctly judged the suspect to be absent. The substantial drop in accuracy for Lineup D indicates that when the substituted non-mate was drawn from the largest (24,000-image) gallery, participants were considerably more likely to be misled into believing the suspect was present, consistent with the increased suspect-to-non-mate similarity reported in Section 7.1.
\begin{figure}[t]
    \centering
    \includegraphics[width=0.85\columnwidth]{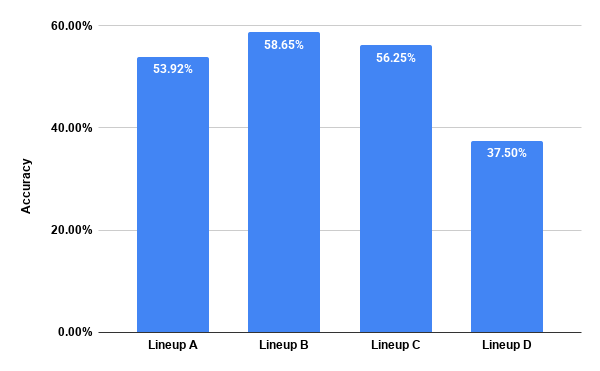}
    \caption{Overall identification accuracy by lineup condition. Lineup A is target-present, where a correct response is selecting the true suspect; Lineups B, C, and D are target-absent, with the suspect replaced by the rank-one non-mate retrieved from the 500-, 5{,}000-, and 24{,}000-image galleries, respectively, where a correct response is judging the suspect to be absent. Accuracy did not differ significantly among Lineups A, B, and C, but Lineup D was significantly less accurate than each of the other three conditions ($\chi^2(3, N = 622) = 13.56$, $p = .004$).}
    \label{fig:accuracy_lineups}
\end{figure}

\subsection{Selection breakdown}
Figure 5 presents the full distribution of participant responses for each lineup condition, categorized as selection of the true suspect (Lineup A only), selection of the rank-one non-mate, selection of a filler image, or no selection (a judgment that the suspect was absent).
In Lineup A, 54.90\% of participants correctly selected the suspect, 14.71\% selected a filler, and 30.39\% incorrectly indicated that no one in the lineup matched their memory of the suspect.
Across the target-absent conditions, the rate at which participants selected the non-mate — mistaking the non-mate for the true suspect — more than doubled at the largest gallery size: 8.17\% in Lineup B and 8.65\% in Lineup C, versus 20.19\% in Lineup D (B vs. D: ($\chi^2(1) = 8.27, p = .004$)); C vs. D: ($\chi^2(1) = 7.42, p = .007$); B vs. C: p = 1.00). Filler selections also rose somewhat with gallery size (32.21\%, 34.13\%, and 42.31\% for B, C, and D, respectively), though these differences did not reach significance $(\text{all } p > 0.10)$. Correspondingly, the rate of correctly judging the suspect to be absent fell from 59.62\% in Lineups B and 57.21\% in Lineup C to 37.50\% in Lineup D.
Together, these results indicate that the drop in overall accuracy observed for Lineup D (Section 7.2) was driven primarily by a sharp increase in the rate at which participants selected the non-mate itself — directly paralleling the rise in suspect-to-non-mate similarity at the 24,000-image gallery size reported in Section 7.1. 
\begin{figure}[t]
    \centering
    \includegraphics[width=0.85\columnwidth]{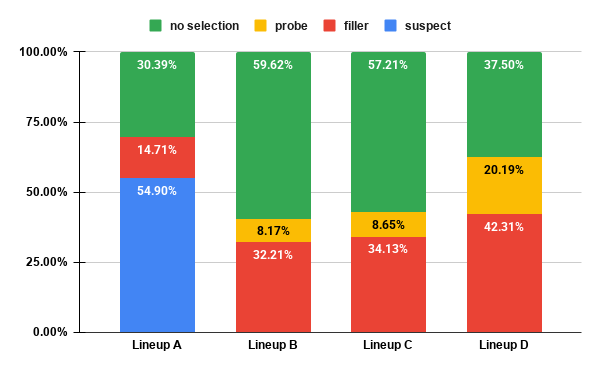}
    \caption{Selection breakdown by lineup condition. Each bar represents the proportion of trials resulting in each response type: selection of the true suspect (blue, Lineup A only), selection of the rank-one non-mate (yellow), selection of a filler (red), or no selection indicating the suspect was judged absent (green). The rate of non-mate selection more than doubled in Lineup D (20.19\%) compared to Lineups B (8.17\%) and C (8.65\%), a significant increase ($\chi^2(1) = 8.27$, $p = .004$) and ($\chi^2(1) = 7.42$, $p = .007$), respectively). No-selection rates fell correspondingly, from approximately 58--60\% in Lineups B and C to 37.50\% in Lineup D.}
    \label{fig:selection_breakdown}
\end{figure}

\subsection{Confidence Scores Across Lineup Conditions}
Confidence ratings were analyzed separately depending on whether participants made a selection (identified someone from the lineup) or made no selection (judged the suspect to be absent). Seven trials were excluded from both analyses as confidence ratings for suspect-absent judgments were not collected.

When a selection was made, mean confidence differed significantly across lineup conditions, $F(3, 305) = 15.84$, $p < .001$. Participants who made a selection in Lineup A (target-present) reported the highest confidence ($M = 7.46$, $SD = 2.08$), significantly higher than those who selected someone in Lineup B ($M = 5.80$, $SD = 2.02$; $t(147.2) = 5.03$, $p < .001$), Lineup C ($M = 5.44$, $SD = 1.92$; $t(144.5) = 6.31$, $p < .001$), and Lineup D ($M = 6.18$, $SD = 1.63$; $t(131.1) = 4.02$, $p < .001$). Among the target-absent conditions, Lineup D produced marginally higher confidence than Lineup C ($t(148.4) = -2.56$, $p = .012$), while Lineups B and C did not differ significantly ($p = .242$), nor did Lineups B and D ($p = .205$).

When no selection was made, confidence ratings were uniformly high and did not differ significantly across conditions ($M = 6.81$, $7.00$, $6.91$, and $6.77$ for Lineups A, B, C, and D respectively; $F(3, 309) = 0.17$, $p = .917$). This indicates that participants who correctly rejected the lineup did so with similar levels of certainty regardless of gallery size.

Taken together, these findings reveal an important asymmetry: while the decision to reject a lineup was made with consistent confidence across conditions, selections made in Lineup D were notably more confident than those in Lineups B and C, despite representing a higher error rate (see Section 7.2). This suggests that the highly similar non-mate retrieved from the 24{,}000-image gallery not only increased the likelihood of misidentification but also induced greater subjective certainty in those erroneous selections --- a particularly concerning combination in the context of eyewitness testimony.

\begin{figure}[t]
    \centering
    \includegraphics[width=0.85\columnwidth]{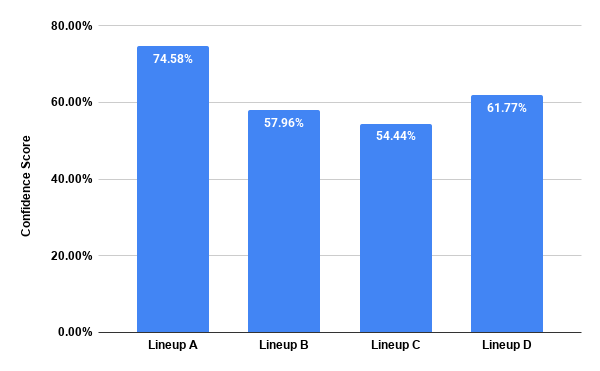}
    \caption{Mean confidence scores (scale 1--10, displayed as percentage of maximum) 
    when a selection was made, by lineup condition. Lineup A participants reported 
    significantly higher confidence than all target-absent conditions 
    ($F(3,305) = 15.84$, $p < .001$). Among target-absent conditions, Lineup D 
    was significantly more confident than Lineup C ($p = .012$).}
    \label{fig:confidence_selection}
\end{figure}

\begin{figure}[t]
    \centering
    \includegraphics[width=0.85\columnwidth]{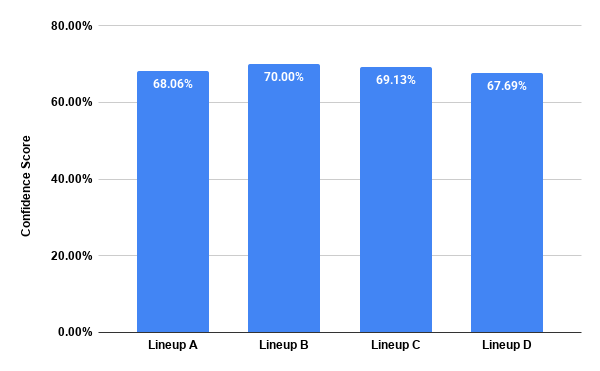}
    \caption{Mean confidence scores (scale 1--10, displayed as percentage of maximum) 
    when no selection was made, by lineup condition. Confidence did not differ 
    significantly across conditions ($F(3,309) = 0.17$, $p = .917$), indicating 
    that lineup rejections were made with similar certainty regardless of gallery size.}
    \label{fig:confidence_no_selection}
\end{figure}

\subsection{Choosing Rates Across Lineup Conditions}
Choosing rate --- the proportion of trials on which a participant selected anyone 
from the lineup rather than judging the suspect to be absent --- differed 
significantly across the four lineup conditions, $\chi^2(3, N = 622) = 34.18$, 
$p < .001$.

Lineup A (target-present) had a choosing rate of 69.61\%, and Lineup D 
(24{,}000-image gallery) had a choosing rate of 62.50\%; these two conditions 
did not differ significantly from one another ($p = .353$). By contrast, 
Lineups B (40.38\%) and C (42.79\%) had substantially lower choosing rates 
that did not differ from each other ($p = .691$) but were each significantly 
lower than both Lineup A (B vs.\ A: $\chi^2(1) = 22.22$, $p < .001$; 
C vs.\ A: $\chi^2(1) = 18.65$, $p < .001$) and Lineup D 
(B vs.\ D: $\chi^2(1) = 12.72$, $p < .001$; 
C vs.\ D: $\chi^2(1) = 10.00$, $p = .002$).

This pattern is notable for two reasons. First, the elevated choosing rate in 
Lineup D relative to Lineups B and C --- despite all three being target-absent 
conditions --- mirrors the rise in suspect-to-non-mate similarity at the 
24{,}000-image gallery size (Section 7.1), suggesting that the more similar 
non-mate drew participants toward making an incorrect identification rather than 
rejecting the lineup. Second, the near-identical choosing rates for Lineup A 
and Lineup D (69.61\% vs.\ 62.50\%, $p = .353$) indicate that participants 
behaved as if a viable suspect were present at roughly the same rate in both 
conditions --- even though Lineup D contained a highly similar non-mate.

\begin{figure}[t]
    \centering
    \includegraphics[width=0.85\columnwidth]{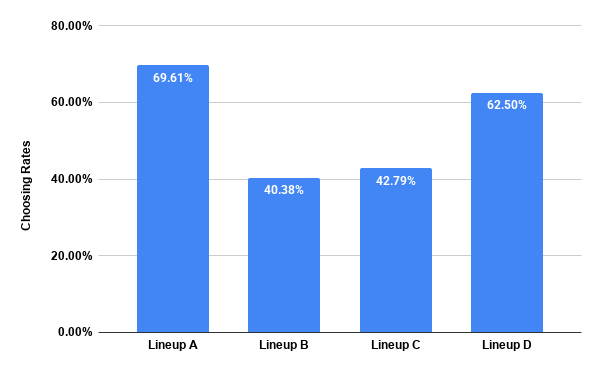}
    \caption{Choosing rates (proportion of trials on which a participant 
    selected anyone from the lineup) by condition. Lineups A and D did not 
    differ significantly ($p = .353$), nor did Lineups B and C ($p = .691$). 
    However, Lineups B and C were each significantly lower than both Lineup A 
    and Lineup D (all $p \leq .002$), indicating that the 24{,}000-image 
    gallery non-mate induced choosing behavior comparable to that seen when 
    the true suspect was present ($\chi^2(3, N = 622) = 34.18$, $p < .001$).}
    \label{fig:choosing_rates}
\end{figure}

\subsection{Frequency and Direction of Response Changes}
In the sequential presentation of lineup images, it is possible that a participant makes an initial selection early in the sequence and then wants to change their selection when seeing a later image. This is noted as a response change. 
Data on response changes were available for 103 of the 104 participants. 
The majority of participants never changed their response during a lineup 
(66 participants, 64.08\%). Among the remaining 37 participants (35.92\%) 
who changed their response at least once, 31 (30.10\%) changed once and 
6 (5.83\%) changed twice, yielding 48 total recorded transitions across 
all lineup conditions.

Of the 48 transitions, the most common was filler-to-filler (f\_to\_f; 
$n = 21$, 43.75\%), reflecting participants who initially selected one 
filler image before switching to a different filler. The second most 
common transition was filler-to-probe (f\_to\_p; $n = 11$, 22.92\%), 
where participants initially selected a filler before settling on the 
rank-one non-mate as their final response. Probe-to-filler transitions 
(p\_to\_f; $n = 8$, 16.67\%) were also observed, indicating participants 
who initially identified the non-mate before revising their selection to 
a filler. Filler-to-suspect transitions (f\_to\_s; $n = 5$, 10.42\%) 
occurred exclusively in Lineup A (target-present), reflecting participants 
who initially selected a filler before correctly identifying the true 
suspect. The remaining transitions were compound: probe-to-filler-to-filler 
(p\_to\_f\_to\_f; $n = 2$, 4.17\%) and filler-to-filler-to-probe 
(f\_to\_f\_to\_p; $n = 1$, 2.08\%).

Considered together, 22 of the 48 transitions (45.83\%) involved the 
rank-one non-mate at some point, with 12 transitions (25.00\%) ending on 
the non-mate as the final selection and 10 transitions (20.83\%) beginning 
on the non-mate before moving away. This suggests that the non-mate image 
was not merely selected in a single committed decision but was actively 
considered and reconsidered during the identification process, consistent 
with the high visual similarity between the non-mate and the true suspect 
established in Section 7.1.

\begin{figure}[t]
    \centering
    \includegraphics[width=0.85\columnwidth]{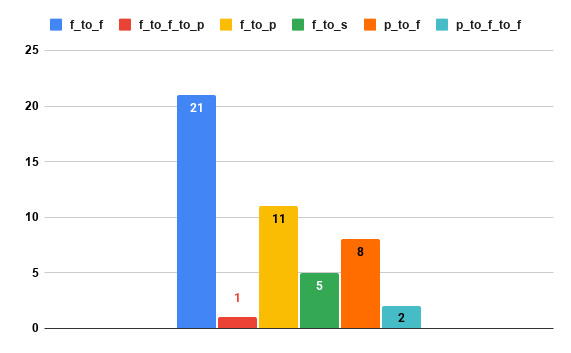}
    \caption{Frequency of each response transition type across all lineup 
    conditions combined ($n = 48$ total transitions from 37 participants). 
    Labels denote transition paths: f = filler, p = rank-one non-mate 
    (probe), s = true suspect. Filler-to-filler transitions were most 
    common (43.75\%), followed by filler-to-probe (22.92\%) and 
    probe-to-filler (16.67\%). Transitions involving the non-mate at 
    any point accounted for 45.83\% of all recorded changes.}
    \label{fig:mind_changes}
\end{figure}

\subsection{Robustness Analysis: Excluding Mind-Changes}
As a secondary analysis, we examined identification accuracy and response distributions restricted to trials on which participants did not change their response during the lineup procedure. This subset reflects protocols used in some jurisdictions in which a witness who revises their identification is considered less reliable or may be excluded from evidentiary consideration \cite{NRC2014}. Accuracy in this restricted sample was 49.44\% for Lineup A, 64.21\% for Lineup B, 58.79\% for Lineup C, and 41.05\% for Lineup D. The pattern of results was consistent with the primary analysis: Lineup D remained the lowest-accuracy condition, and the rate of non-mate selection continued to rise with gallery size (6.73\%, 7.69\%, and 18.27\% for Lineups B, C, and D respectively). These findings suggest that the primary results are robust and not driven by participants who were uncertain enough to revise their decisions mid-procedure.

\section{Conclusions and Discussion}
\label{sec:con&dis}

The overall conclusion from our work is that using the rank-one match from one-to-many facial identification of a large gallery as a suspect image in a photo lineup leads to a higher rate of incorrect identifications. The increased incorrect identifications occur in lineups where the person in the suspect image is not the actual culprit. The larger the gallery that is searched, the greater the similarity between the rank-one non-mate image and the image of the culprit, leading the person viewing the photo lineup to make an incorrect selection of the suspect image rather than a correct rejection of the lineup. Our results suggest that both the frequency of incorrect selections and the confidence in those incorrect selections increase with larger gallery size.

In the standard law enforcement workflow, a trained facial examiner considers the top-N matching images and makes the final selection of a suspect image from these candidates \cite{mcca2021frt}. It seems unlikely that this workflow would affect the overall conclusion. As gallery size increases, the non-mated distribution shifts toward higher similarity scores while the mated distribution remains largely unchanged \cite{pangelinan2024analyzing}, meaning the top-N candidates will all tend to look more like the probe image. Thus, the underlying mechanism that increases the likelihood of incorrect identification remains present even when a human examiner is incorporated into the decision-making process.

The largest gallery size in our experiments is 24,000 images. In the wrongful arrest case of Robert Williams, the probe image from surveillance video in a Detroit retail store was searched against a gallery containing approximately 49 million images \cite{aclu_williams}. Although this database included individuals from all demographic groups and only a subset would be relevant to the search, the example illustrates that real-world facial recognition deployments often involve galleries that are two to three orders of magnitude larger than those used in our experiments. Consequently, while the same qualitative effects observed in our study should be expected in operational settings, the magnitude of those effects may be substantially greater. Specifically, the tendency to incorrectly select the suspect image and to express confidence in that selection is likely to be considerably higher in practice than indicated by our experimental results \cite{aclu_williams}.

In several documented cases of wrongful arrest, including those involving Robert Williams and Porcha Woodruff, the image selected from a one-to-many facial identification search and subsequently used in a photo lineup was an outdated driver's license photograph \cite{aclu_williams,aclu_woodruff}. In these cases, the image used in the lineup represented the individual's appearance at some point in the past rather than their appearance at the time of the investigated incident. More generally, it seems desirable that the image of a suspect used in a photo lineup should be acquired as close in time as possible to the incident under investigation. Other research suggests that if the individual represented by the rank-one gallery image has additional images in the gallery and those images do not also appear near the top of the ranking, this may indicate that the probe image does not correspond to any person present in the gallery \cite{bhatta2025ingallery}. Such information could potentially be used to provide investigators with additional evidence regarding whether a probe image is likely to belong to a gallery member.

These findings raise important concerns regarding the use of images returned by facial recognition systems as investigative leads in criminal investigations. As gallery sizes continue to grow, the probability that a highly similar non-mate image will be returned as a top match also increases. When such images are subsequently incorporated into photo lineups, our results suggest that they systematically bias eyewitness decisions toward incorrect identifications. The combination of increased selection rates and elevated confidence in those selections presents a particular challenge, as confidence is often interpreted as an indicator of identification reliability. Future work should further examine safeguards that can mitigate these effects, and evaluate procedures that reduce the risk of wrongful identification.

\section{Future Work}
\label{sec:future}

This study provides important evidence regarding the effect of database size on human identification accuracy in photo lineups. However, several important directions remain for future investigation.

\textbf{Sequential versus simultaneous lineup presentation.} Our study used sequential lineups, where participants viewed lineup images one at a time rather than all at once, with no opportunity to revisit images. However, law enforcement agencies also use simultaneous lineups, where all images are presented together, and some sequential lineup protocols allow multiple passes through the images. Future work should examine whether database size effects on identification accuracy differ between these presentation methods. Simultaneous lineups may encourage relative judgment strategies and alter decision-making processes \cite{wells1995}, potentially interacting with database size in ways not captured by our sequential design.

\textbf{Demographic diversity and cross-race effects.} Our study focused exclusively on the African-American male cohort from MORPH for selecting suspect images and searching the gallery. This demographic has received the most attention in currently known instances of wrongful arrest \cite{acluWrongfulArrests}. Future research should examine whether the patterns we observed generalize across different demographic groups, including other races, genders, and age ranges. Additionally, collecting participant demographic information would enable investigation of own-race and own-gender bias effects and whether database size interacts with cross-race identification accuracy. This is particularly interesting given contradictory anecdotes about the other-race effect reported by participants in our experiment.

\textbf{Feature matching strategies.} Participants in our study may have employed various strategies when comparing their memory of probe images to lineup photos. Future work should place greater emphasis on understanding specific feature-matching approaches, particularly regarding the relative weight given to hairstyle matching versus facial structure comparison. Explicit instructions or post-task surveys could reveal whether participants over-rely on easily changeable features like hairstyles rather than more stable facial characteristics. Investigating which features drive similarity rankings in the face recognition algorithm, and whether participants' strategies align with these algorithmic priorities, could also provide valuable insights.

\textbf{Learning effects and task familiarity.} Participants may have improved their identification strategies as they gained experience with the lineup task throughout the experiment. For instance, participants might have performed poorly on their first lineup because they initially relied on a single feature (e.g., nose shape), but as they completed more lineups, they may have learned to adopt more comprehensive comparison strategies considering multiple facial features. This would mean that later lineups participant viewed could have systematically different accuracy than their first lineup, regardless of lineup condition. While our experimental design randomized and balanced the assignment of lineup conditions across participants, we did not track the order in which each participant viewed different lineups. Future work should record presentation order to test whether accuracy improves with task experience and whether any learning effects interact with database size conditions.

\textbf{Time pressure and decision confidence.} We did not impose time constraints or collect detailed timing data for participant decisions. On average, participants took three minutes per lineup (range: 1–5 minutes). Future research should examine both response times and decision confidence ratings to better understand the cognitive processes underlying identification decisions. Time pressure may interact with database size effects, potentially forcing participants to rely more heavily on superficial similarities when top-ranked non-matches are highly similar to suspects. Confidence ratings could also reveal whether participants' subjective certainty correlates with actual accuracy across different database sizes.

\textbf{Improved filler image selection.} Our study selected filler images based on overall facial similarity to suspects. However, controlling for specific features—particularly hairstyle similarity—could further isolate the effects of stable facial characteristics. Future work should explore whether ensuring all lineup images share similar hairstyles reduces the tendency to make identification decisions based primarily on this easily changeable feature, and whether this modification interacts with database size effects.

\textbf{Experimental scale and generalizability.} Several methodological enhancements could strengthen future work. A larger number of participants would increase statistical power and enable detection of smaller effect sizes. Similarly, testing a broader range of suspects and lineups would improve generalizability across different facial characteristics. Investigating larger gallery sizes would extend our findings and determine whether the trends we observed continue at database scales more representative of modern face recognition systems. Finally, alternative methods for establishing suspect memory—such as video role-play scenarios with actors, as employed in other eyewitness research \cite{cutler1987}—could provide more ecologically valid conditions that better approximate real-world witness experiences during crimes.

\section{Acknowledgments}
\label{sec:Ack}
We thank Michael Zang for being our guinea pig in testing how long the entire experiment would take. We thank Gabriella Pangelinan for selecting the non-mate images for the 24,000-image gallery. We also thank Aman Bhatta for providing us with the trained ArcFace matcher. Lastly, we thank Professor Karl Ricanek at UNC-W for assembling and popularizing the MORPH dataset.
\clearpage
{
    \small
    \bibliographystyle{ieeenat_fullname}
    \bibliography{main}

@String(CVPR= {IEEE Conf. Comput. Vis. Pattern Recog.})

@String(CVPR  = {CVPR})

@inproceedings{morph,
  author    = {Ricanek, Karl and Tesafaye, Tesfaye Benjamin},
  booktitle = {7th International Conference on Automatic Face and Gesture Recognition (FGR06)},
  title     = {MORPH: A longitudinal facial database},
  year      = {2006},
  pages     = {341--345},
  publisher = {IEEE},
  doi       = {10.1109/FGR.2006.8}
}

@inproceedings{deng2019arcface,
  title={ArcFace: Additive Angular Margin Loss for Deep Face Recognition},
  author={Deng, Jiankang and Guo, Jia and Xue, Niannan and Zafeiriou, Stefanos},
  booktitle={Proceedings of the IEEE/CVF Conference on Computer Vision and Pattern Recognition (CVPR)},
  pages={4690--4699},
  year={2019}
}

@misc{clearview2026overview,
  author = {{Clearview AI}},
  title = {Company Overview},
  year = {2026},
  howpublished = {\url{https://www.clearview.ai/overview}},
  note = {Accessed: 2026-05-23}
}

@misc{acluWrongfulArrests,
  author = {{American Civil Liberties Union}},
  title = {More than a Dozen Wrongful Arrests Due to Police Reliance on Facial Recognition Technology},
  year = {2026},
  url = {https://www.aclu.org/news/privacy-technology/more-than-a-dozen-wrongful-arrests-due-to-police-reliance-on-facial-recognition-technology},
  note = {Accessed: 2026-05-24}
}

@article{loftus2005,
  author = {Loftus, Elizabeth F.},
  title = {Planting misinformation in the human mind: A 30-year investigation of the malleability of memory},
  journal = {Learning \& Memory},
  volume = {12},
  number = {4},
  pages = {361--366},
  year = {2005}
}

@article{wells1995,
  author = {Wells, Gary L. and Seelau, Eric P.},
  title = {Eyewitness Identification: Psychological Research and Legal Policy on Lineups},
  journal = {Psychology, Public Policy, and Law},
  volume = {1},
  number = {4},
  pages = {765--791},
  year = {1995}
}

@article{cutler1987,
  author = {Cutler, Brian L. and Penrod, Steven D. and Martens, Tanja K.},
  title = {Improving the Reliability of Eyewitness Identification: Putting Context into Context},
  journal = {Journal of Applied Psychology},
  volume = {72},
  number = {4},
  pages = {629--637},
  year = {1987}
}

@article{Wells2020,
  author = {Wells, Gary~L. and Kovera, Margaret~Bull and Douglass, Amy~Bradfield and Brewer, Neil and Meissnet, Christian~A. and Wixted, John~T.},
  title = {Policy and procedure recommendations for the collection and preservation of eyewitness identification evidence},
  journal = {Law and Human Behavior},
  volume = {44},
  number = {1},
  pages = {3-36},
  year = {2020}
}

@article{Bergold2018,
  author = {Bergold, Amanda~N. and Heaton, Paul},
  title = {Does Filler Database Size Influence Identification Accuracy?},
  journal = {Law and Human Behavior},
  volume = {42},
  number = {3},
  pages = {227-243},
  year = {2018}
}

@article{Wells2020b,
  author = {Wells, Gary~L.},
  title = {Psychological Science on Eyewitness Identification and Its Impact on Police Practices and Policies},
  journal = {American Psychologist},
  volume = {75},
  number = {9},
  pages = {1316–1329},
  year = {2020}
}

@article{Clothier2024,
  author    = {Clothier, Eden and Michalski, Dana and Malec, Christopher and Nowina-Krowicki, Marcin},
  title     = {Are contemporary facial recognition algorithms making human facial comparison performance worse?},
  journal   = {Forensic Science International},
  volume    = {364},
  pages     = {112202},
  year      = {2024},
  doi       = {10.1016/j.forsciint.2024.112202},
  publisher = {Elsevier BV}
}

@article{Whiteetal2015,
  author    = {White, David and Dunn, James D. and Schmid, Alexandra C. and Kemp, Richard I.},
  title     = {Error Rates in Users of Automatic Face Recognition Software},
  journal   = {PLOS ONE},
  volume    = {10},
  number    = {10},
  pages     = {e0139827},
  year      = {2015},
  doi       = {10.1371/journal.pone.0139827},
  publisher = {Public Library of Science (PLoS)}
}

@article{Kelsoetal2025,
  author    = {Kelso, Lauren E. and Dobolyi, David G. and Grabman, Jesse H. and Dodson, Chad S.},
  title     = {{AI} assistance improves people’s ability to distinguish correct from incorrect eyewitness lineup identifications},
  journal   = {Proceedings of the National Academy of Sciences},
  volume    = {122},
  number    = {21},
  pages     = {e2503971122},
  year      = {2025},
  doi       = {10.1073/pnas.2503971122},
  publisher = {National Academy of Sciences}
}

@inproceedings{johnson2003gallery,
  author    = {Amos Y. Johnson and Jie Sun and Aaron F. Bobick},
  title     = {Using Similarity Scores from a Small Gallery to Estimate Recognition Performance for Larger Galleries},
  booktitle = {Proceedings of the IEEE International Workshop on Analysis and Modeling of Faces and Gestures (AMFG)},
  pages      = {100--103},
  year       = {2003},
  publisher  = {IEEE}
}

@article{friedman2022gallery,
  author  = {Lee Friedman and Hal Stern and Vladyslav Prokopenko and Shagen Djanian and Henry Griffith and Oleg Komogortsev},
  title   = {Biometric Performance as a Function of Gallery Size},
  journal = {Applied Sciences},
  volume  = {12},
  number  = {21},
  pages   = {11144},
  year    = {2022},
  doi     = {10.3390/app122111144}
}

@inproceedings{vareto2016largegalleries,
  author    = {Rafael H. Vareto and Filipe de O. Costa and William Robson Schwartz},
  title     = {Face Identification in Large Galleries},
  booktitle = {Workshop on Face Processing Applications (WFPA) in the 29th Conference on Graphics, Patterns and Images (SIBGRAPI)},
  year      = {2016}
}

@misc{Indiana2025,
  title={SENATE ENROLLED ACT No. 141 / Public Law 127},
  author={State of Indiana},
  year={2025},
  howpublished = {\url{https://legiscan.com/IN/text/SB0141/id/3213933}}
}

@misc{mcca2021frt,
  title={Facial Recognition Technology in Modern Policing: Recommendations and Considerations},
  author={{Major Cities Chiefs Association Facial Recognition Working Group}},
  year={2021},
  howpublished = {\url{https://majorcitieschiefs.com/wp-content/uploads/2021/10/MCCA-FRT-in-Modern-Policing-Final.pdf}}
}

@misc{aclu_williams,
  title={Williams v. City of Detroit},
  author={{American Civil Liberties Union}},
  year={2024},
  howpublished = {\url{https://www.aclu.org/cases/williams-v-city-of-detroit-face-recognition-false-arrest}}
}

@misc{aclu_woodruff,
  title={Porcha Woodruff Wrongful Arrest Case},
  author={{American Civil Liberties Union}},
  year={2023}
}

@inproceedings{bhatta2025ingallery,
  title={Are You In or Out (of Gallery)? Wisdom from the Same-Identity Crowd},
  author={Bhatta, Aman and Dhakal, Maria and King, Michael C. and Bowyer, Kevin W.},
  booktitle={Proceedings of the IEEE/CVF International Conference on Computer Vision},
  pages={5946--5955},
  year={2025}
}

@misc{Richardson_article,
  author = {Brand, Eli},
  title = {{AI} misidentification results in wrongful arrest; man seeks justice},
  journal = {{WSOC-TV}},
  year = {2026},
  month = {June},
  day = {11},
  howpublished = {\url{https://www.wsoctv.com/news/local/ai-misidentification-results-wrongful-arrest-man-seeks-justice/I7UQJWV33FBN3LMKHCSXI6FIVA/}}
}

@misc{Burgess_article,
  author = {Schecker, Justin},
  title = {Central Florida man wrongly arrested after Orlando police used facial recognition},
  journal = {WESH 2 News},
  year = {2026},
  month = {February},
  day = {18},
    howpublished = {\url{https://www.wesh.com/article/florida-man-wrongly-arrested-police-facial-recognition/70409653}}

}

@article{Woodruff_article,
  author = {{Associated Press}},
  title = {Woman wrongly accused of carjacking loses lawsuit against Detroit police who used facial technology},
  journal = {CBS News Detroit},
  year = {2025},
  month = {September},
  day = {4}
}

@misc{Dillon_article,
  author = {{ACLU of Florida}},
  title = {Florida Man Sues Police Over Wrongful Arrest Due to False Facial Recognition Match},
  year = {2026},
  month = {June},
  day = {10},
  howpublished = {\url{https://www.aclu.org/press-releases/florida-man-sues-police-over-wrongful-arrest-due-to-false-facial-recognition-match}}
}

@article{T_Williams_article,
  author = {Tarita, Tudor},
  title = {NYC Man Was Jailed for Days Because of a Blurry CCTV Image and a Faulty AI Match},
  journal = {ZME Science},
  year = {2025},
  month = {September},
  day = {7},
  howpublished = {\url{https://www.zmescience.com/science/news-science/nyc-man-was-jailed-for-days-because-of-a-blurry-cctv-image-and-a-faulty-ai-match/}}  
}

@misc{Oliver_article,
  author = {Anderson, Elisha},
  title = {Controversial Detroit facial recognition got him arrested for a crime he didn't commit},
  journal = {Detroit Free Press},
  year = {2020},
  month = {July},
  day = {10},
  howpublished = {\url{https://www.freep.com/story/news/local/michigan/detroit/2020/07/10/facial-recognition-detroit-michael-oliver-robert-williams/5392166002/}}
}

@article{Sawyer_article,
  author = {Press, Eyal},
  title = {Does A.I. Lead Police to Ignore Contradictory Evidence?},
  journal = {The New Yorker},
  year = {2023},
  month = {November},
  day = {13}
}

@misc{Gatlin_article,
  author = {Hayes, Chris},
  title = {Man jailed over police AI program, then freed 17 months after victim raised doubts},
  journal = {FOX8 News},
  year = {2025},
  month = {February},
  day = {5},
    howpublished = {\url{https://fox2now.com/news/fox-files/jailed-over-police-ai-program-then-freed-17-months-after-victim-raised-doubts/}}
}

@misc{FRVT_Identification,
  author       = {Grother, Patrick and Ngan, Mei and Hanaoka, Kayee},
    title = {{Face Recognition Vendor Test ({FRVT}) Part 2: Identification}},
  institution  = {National Institute of Standards and Technology (NIST)},
  series       = {NIST Interagency/Internal Report (NISTIR)},
  number       = {8271},
  year         = {2019},
  doi          = {10.6028/NIST.IR.8271},
  howpublished = {\url{https://doi.org/10.6028/NIST.IR.8271}}
}

@book{NRC2014,
  author    = {{National Research Council}},
  title     = {Identifying the Culprit: Assessing Eyewitness Identification},
  publisher = {The National Academies Press},
  address   = {Washington, DC},
  year      = {2014},
  doi       = {10.17226/18891},
  url       = {https://doi.org/10.17226/18891}
}

@INPROCEEDINGS{uncw_morph,
  author={Ricanek, K. and Tesafaye, T.},
  booktitle={7th International Conference on Automatic Face and Gesture Recognition (FGR06)},
  title={MORPH: a longitudinal image database of normal adult age-progression},
  year={2006}
}

@techreport{wisconsin_eyewitness_2026,
  author      = {{Wisconsin Department of Justice}},
  institution = {Bureau of Training and Standards for Criminal Justice},
  title       = {State of {Wisconsin} Model Policy and Procedure for Eyewitness Identification},
  year        = {2026},
  type        = {Model Policy and Procedure},
  address     = {Madison, WI},
  note        = {Office of the Attorney General},
  url         = {https://vodgsearch.org/Eyewitness/WisconsinAGEyewitnessIdentificationGuidelines.pdf}
}

@article{MeissnerBrigham2001,
  author    = {Meissner, Christian A. and Brigham, John C.},
  title     = {Thirty Years of Investigating the Own-Race Bias in Memory 
               for Faces: A Meta-Analytic Review},
  journal   = {Psychology, Public Policy, and Law},
  volume    = {7},
  number    = {1},
  pages     = {3--35},
  year      = {2001},
  doi       = {10.1037/1076-8971.7.1.3}
}

@misc{Woodruff_court,
  author       = {{Law Offices of Ivan L. Land, P.C.}},
  title        = {Woodruff v. City of Detroit, Complaint and Jury Demand},
  year         = {2023},
  howpublished = {\url{https://www.courtlistener.com/docket/67661093/woodruff-v-detroit-city-of/}},
  note         = {Accessed: 2026-06-25}
}

@incollection{Fitzgerald2021,
  author    = {Fitzgerald, Ryan J. and Rub{\'i}nov{\'a}, Eva and Juncu, Stefana},
  title     = {Eyewitness Identification Around the World},
  booktitle = {Methods, Measures, and Theories in Eyewitness Identification Tasks},
  edition   = {1},
  publisher = {Routledge},
  year      = {2021},
  pages     = {29},
  doi       = {10.4324/9781003138105}
}

@article{pangelinan2024analyzing,
  author    = {Pangelinan, Gabriella and Bhatta, Aman and Wu, Haiyu and King, Michael C. and Bowyer, Kevin W.},
  title     = {Analyzing the Impact of Demographic and Operational Variables on 1-to-Many Face {ID} Search},
  journal   = {IEEE Transactions on Technology and Society},
  volume    = {5},
  number    = {2},
  pages     = {217--230},
  year      = {2024},
  publisher = {IEEE},
  doi       = {10.1109/TTS.2024.3425350}
}

@misc{mcda2015eyewitness,
  author       = {{Multnomah County District Attorney's Office}},
  title        = {Eyewitness Identification Form: Photo Lineup},
  year         = {2015},
  month        = aug,
  howpublished = {Official Policy and Forms Document},
  address      = {Portland, OR},
  url          = {https://www.mcda.us/index.php/documents/eyewitness-identification-forms.pdf},
  note         = {Accessed: July 2026}
}

@misc{R_Williams_article,
  author       = {Williams, Robert},
  title        = {I Was Wrongfully Arrested Because of Facial Recognition Technology. 
                  It Shouldn't Happen to Anyone Else},
  year         = {2024},
  month        = {July},
  howpublished = {\url{https://www.aclumich.org/news/i-was-wrongfully-arrested-because-facial-recognition-technology-it-shouldnt-happen-anyone-else/}},
  note         = {Accessed: 2026-07-08}
}

@techreport{gao2019face,
  author      = {{U.S. Government Accountability Office}},
  title       = {Face Recognition Technology: {DOJ} and {FBI} Have Taken Some Actions in Response to {GAO} Recommendations to Ensure Privacy and Accuracy, But Additional Work Remains}, 
  institution = {U.S. Government Accountability Office},
  address     = {Washington, D.C.},
  number      = {GAO-19-579T},
  year        = {2019},
  month       = {June},
  note        = {Testimony before the Committee on Oversight and Reform, House of Representatives},
  url         = {https://www.gao.gov/products/gao-19-579t}
}

@article{meyer2023interactive,
  author    = {Meyer, M. and Colloff, M.F. and Bennett, T.C. and Hirata, E. 
               and Kohl, A. and Stevens, L.M. and Smith, H.M.J. and 
               Staudigl, T. and Flowe, H.D.},
  title     = {Enabling witnesses to actively explore faces and reinstate 
               study-test pose during a lineup increases discriminability},
  journal   = {Proceedings of the National Academy of Sciences},
  volume    = {120},
  number    = {41},
  pages     = {e2301845120},
  year      = {2023},
  doi       = {10.1073/pnas.2301845120}
}

@article{Heyer2018,
  author    = {Heyer, Rebecca and Semmler, Carolyn and  Hendrickson, Andrew~T.},
  title     = {Humans and Algorithms for Facial Recognition: The Effects of Candidate List Length and Experience on Performance},
  journal   = {Journal of Applied Research in Memory and Cognition},
  volume    = {7},
  pages     = {597-609},
  year      = {2018}
}

@article{fitzgerald2025eli,
  author    = {Fitzgerald, Ryan J. and Rubínová, Eva and Ribbers, Eva 
               and Juncu, Stefana},
  title     = {Eyewitness Lineup Identity ({ELI}) database: Crime videos 
               and mugshots for eyewitness identification research},
  journal   = {Behavior Research Methods},
  volume    = {57},
  pages     = {63},
  year      = {2025},
  doi       = {10.3758/s13428-024-02585-z}
}

@article{Albiero_2022,
   title={Gendered Differences in Face Recognition Accuracy Explained by Hairstyles, Makeup, and Facial Morphology},
   volume={17},
   ISSN={1556-6021},
   url={http://dx.doi.org/10.1109/TIFS.2021.3135750},
   DOI={10.1109/tifs.2021.3135750},
   journal={IEEE Transactions on Information Forensics and Security},
   publisher={Institute of Electrical and Electronics Engineers (IEEE)},
   author={Albiero, Vitor and Zhang, Kai and King, Michael C. and Bowyer, Kevin W.},
   year={2022},
   pages={127–137} }
}
\clearpage
\onecolumn

\label{sec:datasetUsed}

\begin{figure*}[h!] 
\centering
\renewcommand{\arraystretch}{1.2} 
\setlength{\tabcolsep}{2pt}       

\begin{tabular}{ccccc}
\fbox{\parbox{0.15\textwidth}{\centering \includegraphics[width=\linewidth]{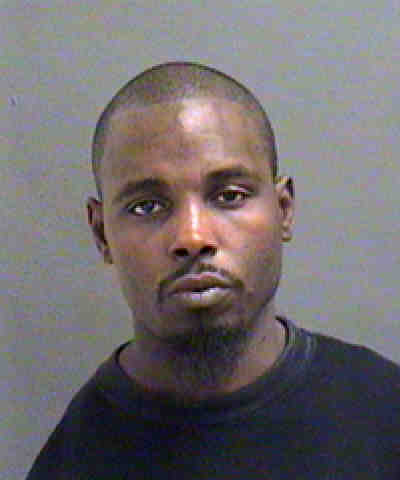} \\ \tiny Probe image \\ 204370\_04M32.JPG}} &
\fbox{\parbox{0.15\textwidth}{\centering \includegraphics[width=\linewidth]{figs/experiment1/lineup1/204370_05M32.JPG} \\ \tiny Mated gallery image \\ 204370\_05M32.JPG}} &
\fbox{\parbox{0.15\textwidth}{\centering \includegraphics[width=\linewidth]{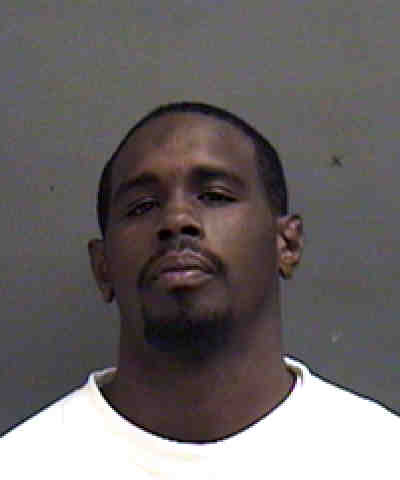} \\ \tiny Rank-one non-mate\\ from gallery of 500 \\ 341043\_00M25.JPG}} &
\fbox{\parbox{0.15\textwidth}{\centering \includegraphics[width=\linewidth]{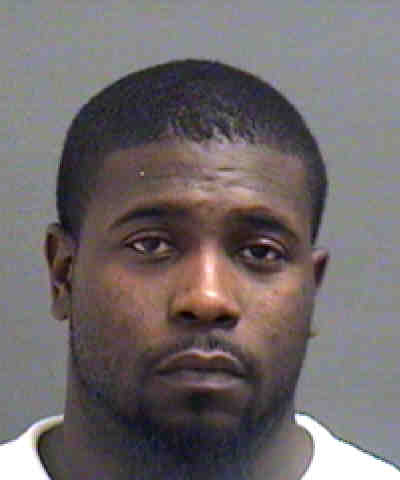} \\ \tiny Rank-one non-mate\\ from gallery of 5,000 \\ 182093\_00M28.JPG}} &
\fbox{\parbox{0.15\textwidth}{\centering \includegraphics[width=\linewidth]{figs/experiment1/lineup4/probe_204370_rank1_206987_fullsize.JPG} \\ \tiny Rank-one non-mate\\ from gallery of 24,000 \\ 11206987\_04M29.JPG}} \\

\fbox{\parbox{0.15\textwidth}{\centering \includegraphics[width=\linewidth]{figs/experiment1/lineup4/068291_03M50.JPG} \\ \tiny Filler image \#1 \\ 068291\_03M50.JPG}} &
\fbox{\parbox{0.15\textwidth}{\centering \includegraphics[width=\linewidth]{figs/experiment1/lineup4/114513_06M48.JPG} \\ \tiny Filler image \#2 \\ 114513\_06M48.JPG}} &
\fbox{\parbox{0.15\textwidth}{\centering \includegraphics[width=\linewidth]{figs/experiment1/lineup4/161494_08M36.JPG} \\ \tiny Filler image \#3 \\ 161494\_08M36.JPG}} &
\fbox{\parbox{0.15\textwidth}{\centering \includegraphics[width=\linewidth]{figs/experiment1/lineup4/231716_07M27.JPG} \\ \tiny Filler image \#4 \\ 231716\_07M27.JPG}} &
\fbox{\parbox{0.15\textwidth}{\centering \includegraphics[width=\linewidth]{figs/experiment1/lineup4/259325_01M35.JPG} \\ \tiny Filler image \#5 \\ 259325\_01M35.JPG}} \\

\fbox{\parbox{0.15\textwidth}{\centering \includegraphics[width=\linewidth]{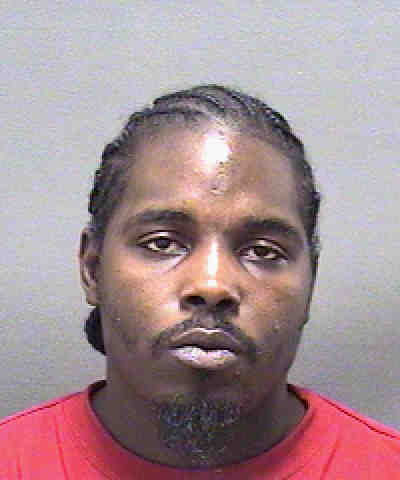} \\ \tiny Memory acquisition task\\ mated image \#1 \\ 204370\_00M28.JPG}} &
\fbox{\parbox{0.15\textwidth}{\centering \includegraphics[width=\linewidth]{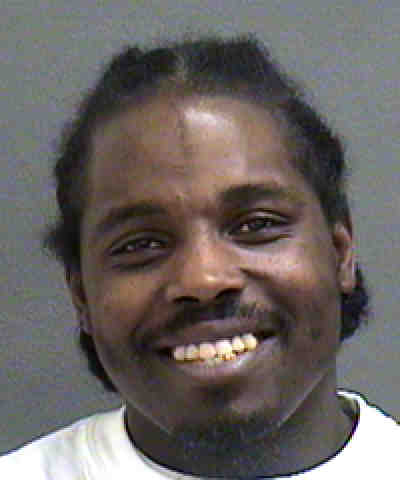} \\ \tiny Memory acquisition task\\ mated image \#2 \\ 204370\_01M28.JPG}} &
\fbox{\parbox{0.15\textwidth}{\centering \includegraphics[width=\linewidth]{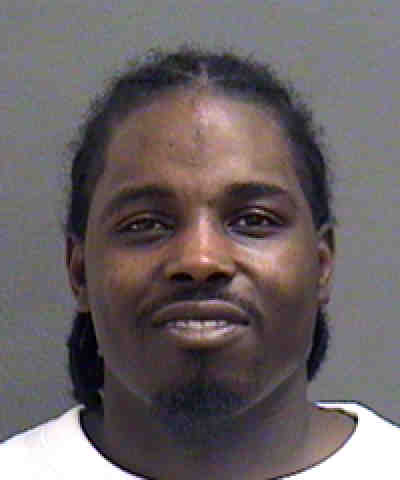} \\ \tiny Memory acquisition task\\ mated image \#3 \\ 204370\_02M28.JPG}} &
\fbox{\parbox{0.15\textwidth}{\centering \includegraphics[width=\linewidth]{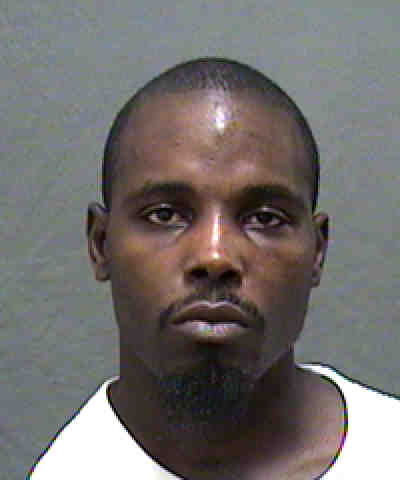} \\ \tiny Memory acquisition task\\ mated image \#4 \\ 204370\_03M32.JPG}} &
\\

\fbox{\parbox{0.15\textwidth}{\centering \includegraphics[width=\linewidth]{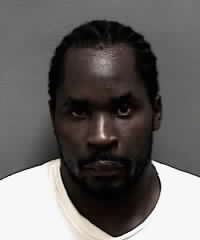} \\ \tiny Memory acquisition task\\ non-mated image \#1 \\ 097390\_01M33.JPG}} &
\fbox{\parbox{0.15\textwidth}{\centering \includegraphics[width=\linewidth]{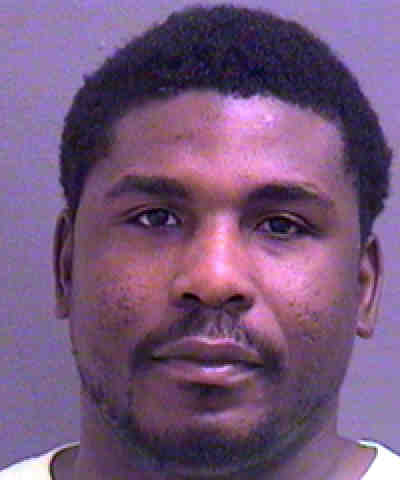} \\ \tiny Memory acquisition task\\ non-mated image \#2 \\ 174871\_06M32.JPG}} &
\fbox{\parbox{0.15\textwidth}{\centering \includegraphics[width=\linewidth]{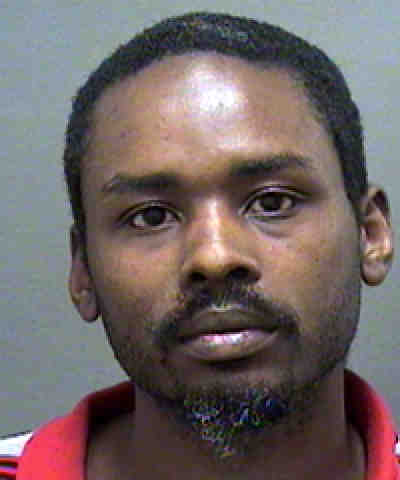} \\ \tiny Memory acquisition task\\ non-mated image \#3 \\ 180464\_01M33.JPG}} &
\fbox{\parbox{0.15\textwidth}{\centering \includegraphics[width=\linewidth]{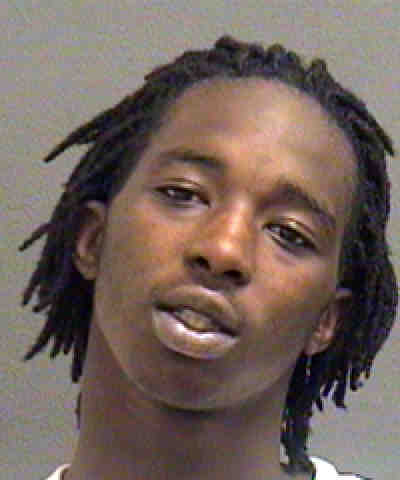} \\ \tiny Memory acquisition task\\ non-mated image \#4 \\ 313263\_04M18.JPG}} &
\\
\end{tabular}

\caption{"Suspect 1" images used with six-pack photo lineups in experiment. Top row: image of suspect 1 used as probe to search against galleries, mated image for target present gallery, and rank-one non-mated images from galleries of varying sizes. Second row: filler images. Rows 3 and 4: images used in face memory acquisition task.}

\label{fig:suspect1} 
\end{figure*}

\begin{figure*}[t] 
\centering
\renewcommand{\arraystretch}{1.8} 
\setlength{\tabcolsep}{2pt}       

\begin{tabular}{ccccc}
\fbox{\parbox{0.17\textwidth}{\centering \includegraphics[width=\linewidth]{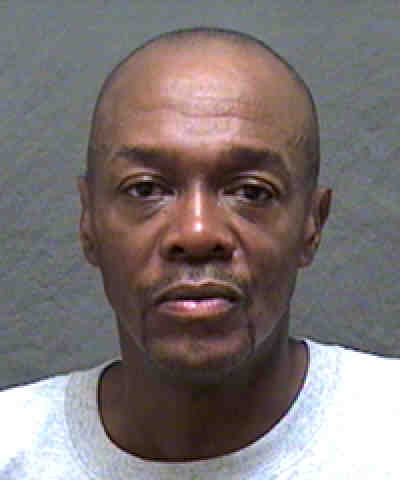} \\ \footnotesize Probe image \\ 061018\_06M51.JPG}} &
\fbox{\parbox{0.17\textwidth}{\centering \includegraphics[width=\linewidth]{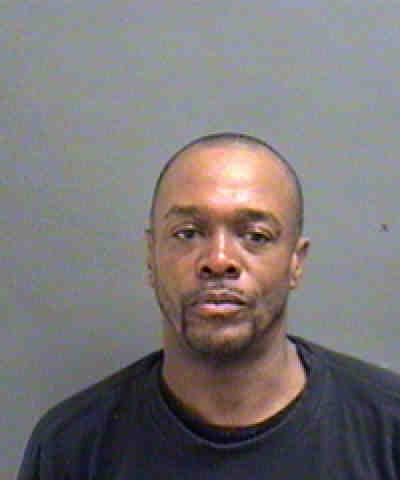} \\ \footnotesize Mated\\ gallery image \\ 061018\_00M47.JPG}} &
\fbox{\parbox{0.17\textwidth}{\centering \includegraphics[width=\linewidth]{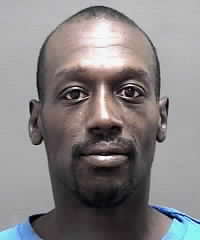} \\ \footnotesize Rank-one non-mate\\ from gallery of 500 \\ 071712\_00M36.JPG}} &
\fbox{\parbox{0.17\textwidth}{\centering \includegraphics[width=\linewidth]{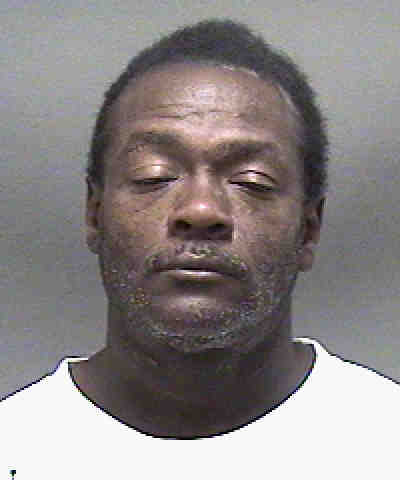} \\ \footnotesize Rank-one non-mate\\ from gallery of 5,000 \\ 286968\_08M44.JPG}} &
\fbox{\parbox{0.17\textwidth}{\centering \includegraphics[width=\linewidth]{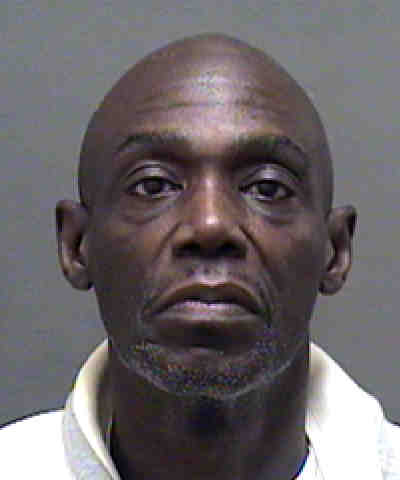} \\ \footnotesize Rank-one non-mate\\ from gallery of 24,000 \\ 11196750\_04M57}} \\

\fbox{\parbox{0.17\textwidth}{\centering \includegraphics[width=\linewidth]{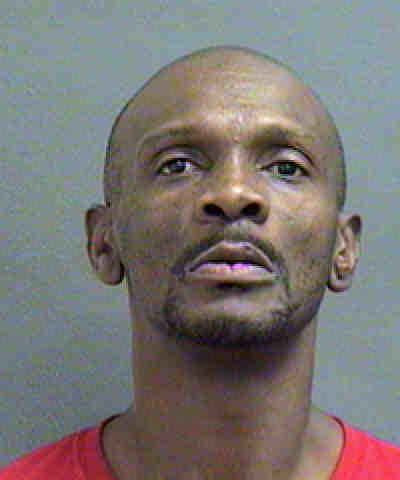} \\ \footnotesize Filler image \#1 \\ 062530\_16M44.JPG}} &
\fbox{\parbox{0.17\textwidth}{\centering \includegraphics[width=\linewidth]{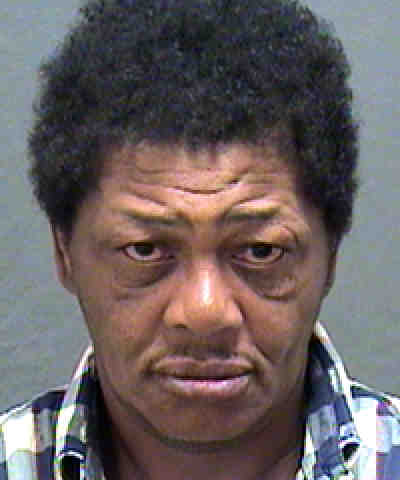} \\ \footnotesize Filler image \#2 \\ 075618\_16M54.JPG}} &
\fbox{\parbox{0.17\textwidth}{\centering \includegraphics[width=\linewidth]{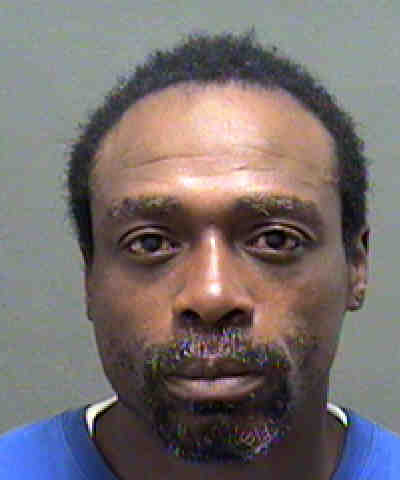} \\ \footnotesize Filler image \#3 \\ 130376\_12M44.JPG}} &
\fbox{\parbox{0.17\textwidth}{\centering \includegraphics[width=\linewidth]{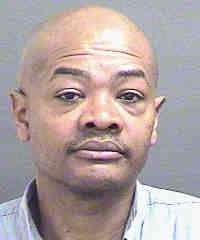} \\ \footnotesize Filler image \#4 \\ 135939\_02M40.JPG}} &
\fbox{\parbox{0.17\textwidth}{\centering \includegraphics[width=\linewidth]{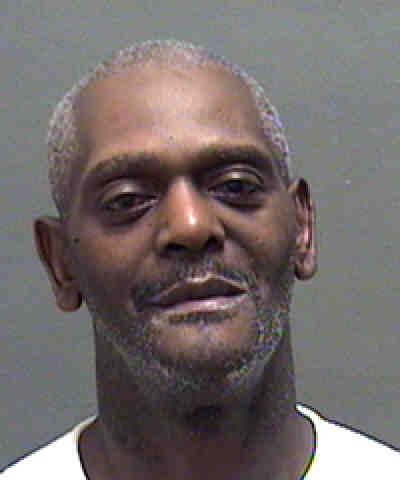} \\ \footnotesize Filler image \#5 \\ 334302\_02M53.JPG}} \\

\fbox{\parbox{0.17\textwidth}{\centering \includegraphics[width=\linewidth]{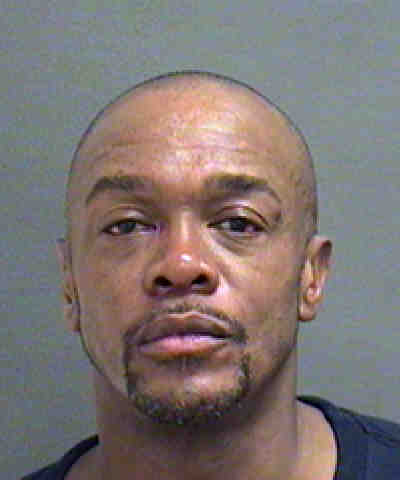} \\ \footnotesize Memory acquisition task\\ mated image \#1 \\ 061018\_01M47.JPG}} &
\fbox{\parbox{0.17\textwidth}{\centering \includegraphics[width=\linewidth]{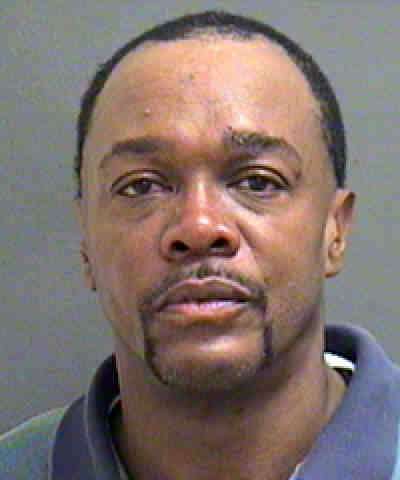} \\ \footnotesize Memory acquisition task\\ mated image \#2 \\ 061018\_03M48.JPG}} &
\fbox{\parbox{0.17\textwidth}{\centering \includegraphics[width=\linewidth]{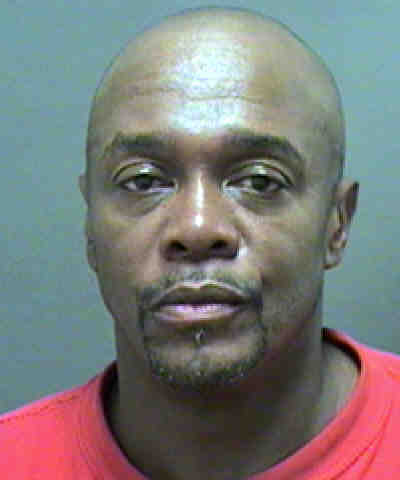} \\ \footnotesize memory acquisition task\\ mated image \#3 \\ 061018\_04M49.JPG}} &
\fbox{\parbox{0.17\textwidth}{\centering \includegraphics[width=\linewidth]{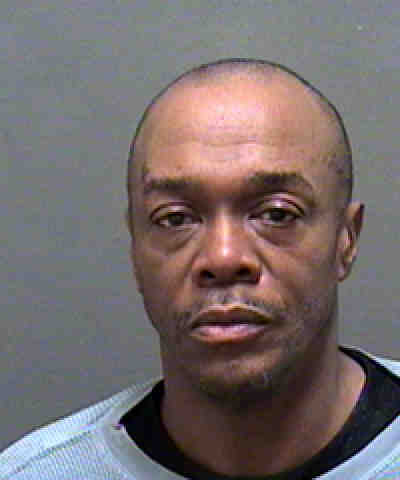} \\ \footnotesize memory acquisition task\\ mated image \#4 \\ 061018\_05M50.JPG}} &
\\

\fbox{\parbox{0.17\textwidth}{\centering \includegraphics[width=\linewidth]{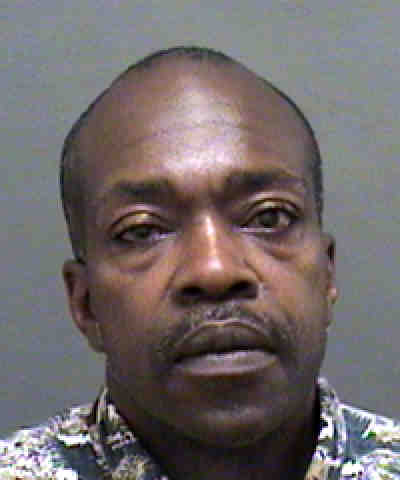} \\ \footnotesize Memory acquisition task\\ non-mated image \#1 \\ 054533\_09M50.JPG}} &
\fbox{\parbox{0.17\textwidth}{\centering \includegraphics[width=\linewidth]{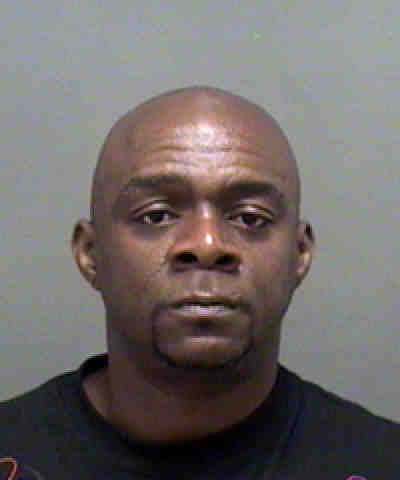} \\ \footnotesize Memory acquisition task\\ non-mated image \#2 \\ 089711\_15M44.JPG}} &
\fbox{\parbox{0.17\textwidth}{\centering \includegraphics[width=\linewidth]{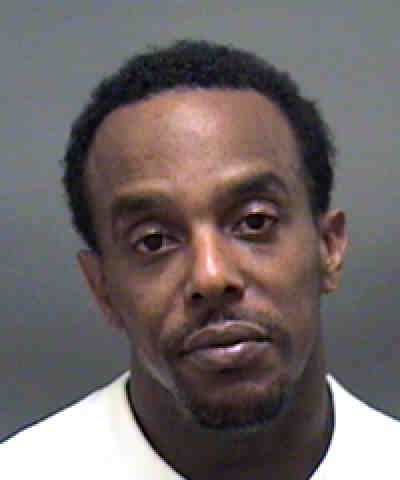} \\ \footnotesize Memory acquisition task\\ non-mated image \#3 \\ 254565\_02M38.JPG}} &
\fbox{\parbox{0.17\textwidth}{\centering \includegraphics[width=\linewidth]{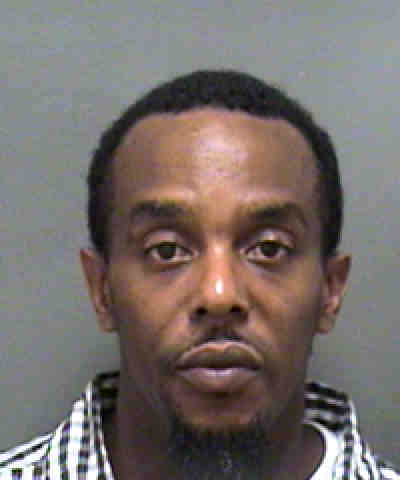} \\ \footnotesize Memory acquisition task\\ non-mated image \#4 \\ 254565\_03M39.JPG}} &
\\
\end{tabular}

\caption{“Suspect 2” images used with six-pack photo lineups in experiment. Top row: image of suspect 1 used as probe to search against galleries, mated image for target present gallery, and rank-one non-mated images from galleries of varying sizes. Second row: filler images. Rows 3 and 4: images used in face memory acquisition task.}

\label{fig:suspect2} 
\end{figure*}

\begin{figure*}[t] 
\centering
\renewcommand{\arraystretch}{1.8} 
\setlength{\tabcolsep}{2pt}       

\begin{tabular}{ccccc}
\fbox{\parbox{0.17\textwidth}{\centering \includegraphics[width=\linewidth]{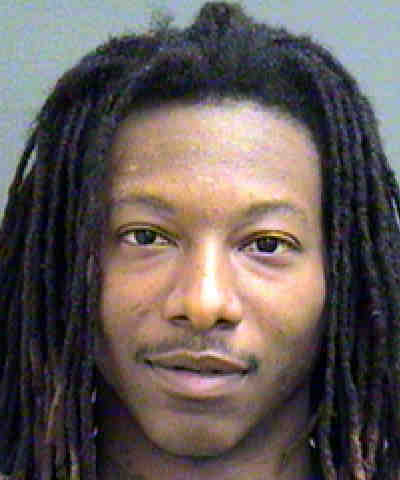} \\ \footnotesize Probe image \\ 361054\_08M22.JPG}} &
\fbox{\parbox{0.17\textwidth}{\centering \includegraphics[width=\linewidth]{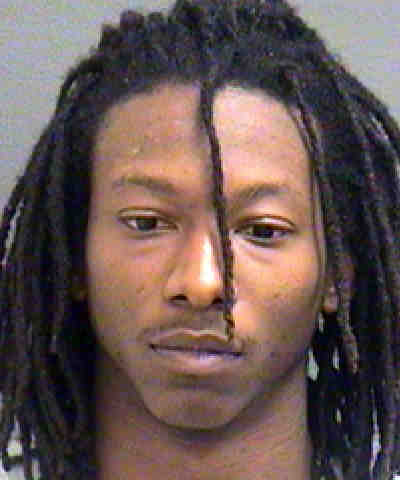} \\ \footnotesize Mated\\ gallery image \\ 361054\_03M20.JPG}} &
\fbox{\parbox{0.17\textwidth}{\centering \includegraphics[width=\linewidth]{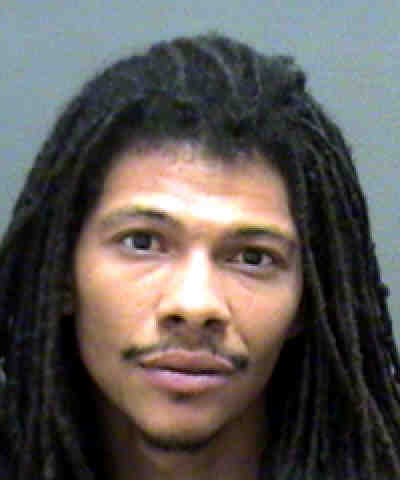} \\ \footnotesize Rank-one non-mate\\ from gallery of 500 \\ 224004\_11M29.JPG}} &
\fbox{\parbox{0.17\textwidth}{\centering \includegraphics[width=\linewidth]{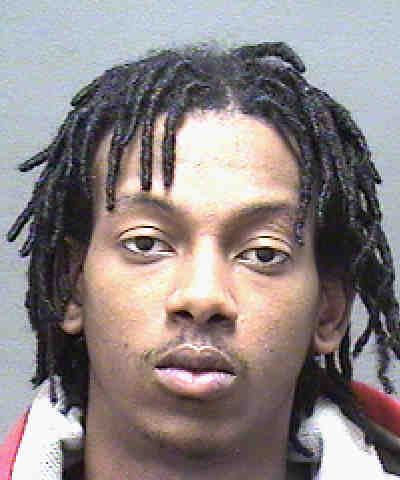} \\ \footnotesize Rank-one non-mate\\ from gallery of 5,000 \\ 295277\_02M20.JPG}} \\

\fbox{\parbox{0.17\textwidth}{\centering \includegraphics[width=\linewidth]{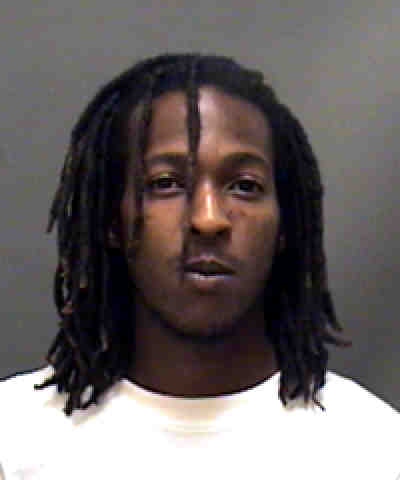} \\ \footnotesize Filler image \#2 \\ 290119\_01M22.JPG}} &
\fbox{\parbox{0.17\textwidth}{\centering \includegraphics[width=\linewidth]{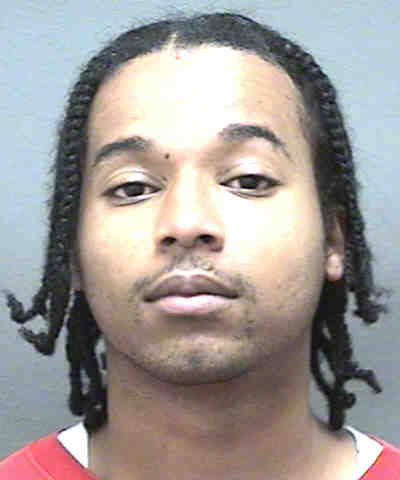} \\ \footnotesize Filler image \#2 \\ 294649\_00M21.JPG}} &
\fbox{\parbox{0.17\textwidth}{\centering \includegraphics[width=\linewidth]{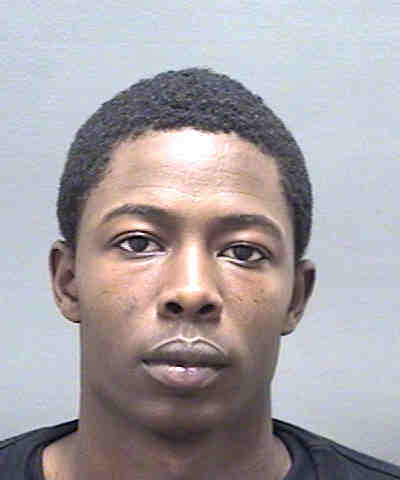} \\ \footnotesize Filler image \#3 \\ 310153\_02M19.JPG}} &
\fbox{\parbox{0.17\textwidth}{\centering \includegraphics[width=\linewidth]{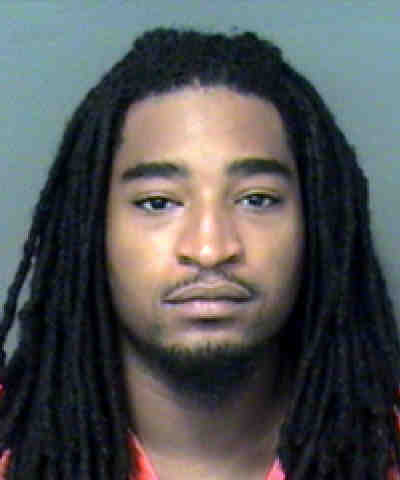} \\ \footnotesize Filler image \#4 \\ 315029\_13M23.JPG}} &
\fbox{\parbox{0.17\textwidth}{\centering \includegraphics[width=\linewidth]{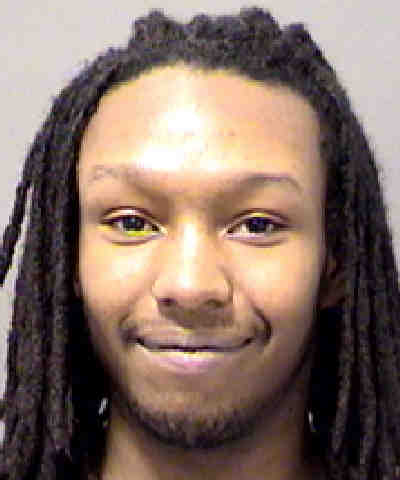} \\ \footnotesize Filler image \#5 \\ 373896\_03M21.JPG}} \\

\fbox{\parbox{0.17\textwidth}{\centering \includegraphics[width=\linewidth]{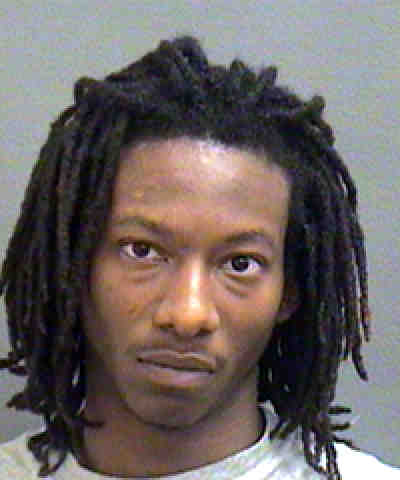} \\ \footnotesize Memory acquisition task\\ mated image \#1 \\ 361054\_01M20.JPG}} &
\fbox{\parbox{0.17\textwidth}{\centering \includegraphics[width=\linewidth]{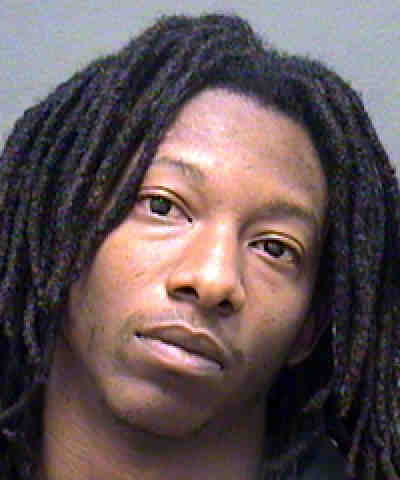} \\ \footnotesize Memory acquisition task\\ mated image \#2 \\ 361054\_04M20.JPG}} &
\fbox{\parbox{0.17\textwidth}{\centering \includegraphics[width=\linewidth]{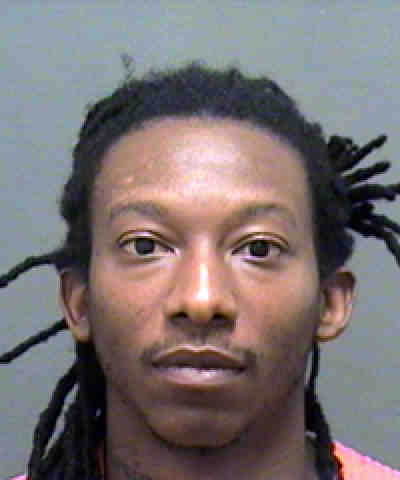} \\ \footnotesize memory acquisition task\\ mated image \#3 \\ 361054\_11M22.JPG}} &
\fbox{\parbox{0.17\textwidth}{\centering \includegraphics[width=\linewidth]{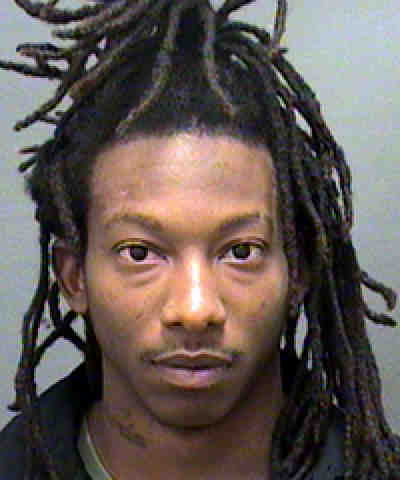} \\ \footnotesize memory acquisition task\\ mated image \#4 \\ 361054\_06M21.JPG}} &
\\

\fbox{\parbox{0.17\textwidth}{\centering \includegraphics[width=\linewidth]{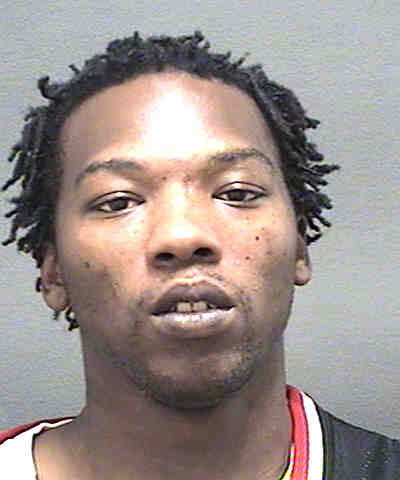} \\ \footnotesize Memory acquisition task\\ non-mated image \#1 \\ 184512\_00M27.JPG}} &
\fbox{\parbox{0.17\textwidth}{\centering \includegraphics[width=\linewidth]{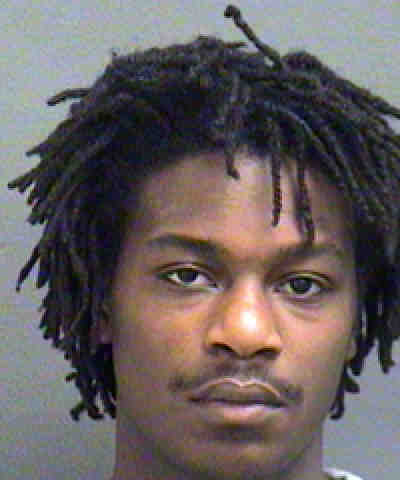} \\ \footnotesize Memory acquisition task\\ non-mated image \#2 \\ 340006\_10M19.JPG}} &
\fbox{\parbox{0.17\textwidth}{\centering \includegraphics[width=\linewidth]{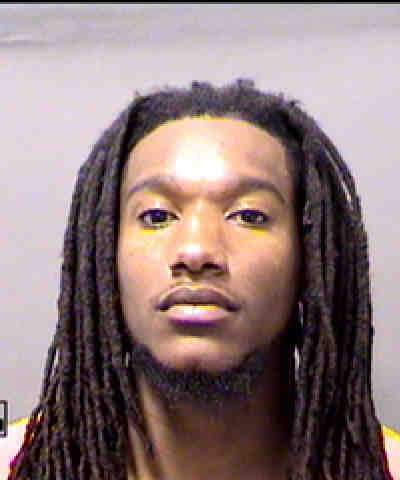} \\ \footnotesize Memory acquisition task\\ non-mated image \#3 \\ 340356\_05M22.JPG}} &
\fbox{\parbox{0.17\textwidth}{\centering \includegraphics[width=\linewidth]{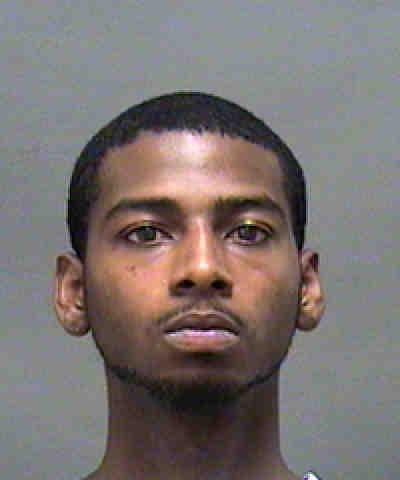} \\ \footnotesize Memory acquisition task\\ non-mated image \#4 \\ 361992\_01M21.JPG}} &
\\
\end{tabular}

\caption{“Suspect 3” images used with six-pack photo lineups in experiment. Top row: image of suspect 1 used as probe to search against galleries, mated image for target present gallery, and rank-one non-mated images from galleries of varying sizes. Second row: filler images. Rows 3 and 4: images used in face memory acquisition task.}

\label{fig:suspect3} 
\end{figure*}

\begin{figure*}[t] 
\centering
\renewcommand{\arraystretch}{1.8} 
\setlength{\tabcolsep}{2pt}       

\begin{tabular}{ccccc}
\fbox{\parbox{0.17\textwidth}{\centering \includegraphics[width=\linewidth]{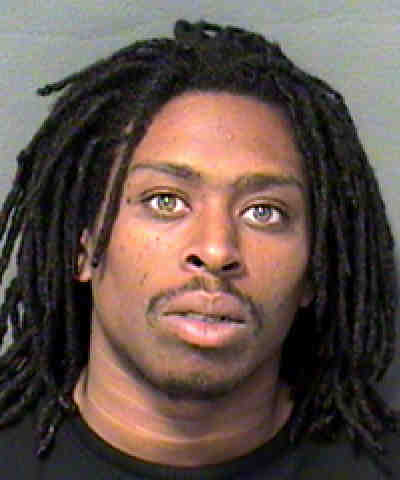} \\ \footnotesize Probe image \\ 259095\_08M28.JPG}} &
\fbox{\parbox{0.17\textwidth}{\centering \includegraphics[width=\linewidth]{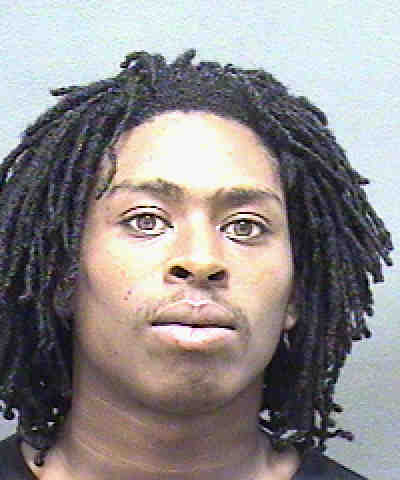} \\ \footnotesize Mated\\ gallery image \\ 259095\_04M23.JPG}} &
\fbox{\parbox{0.17\textwidth}{\centering \includegraphics[width=\linewidth]{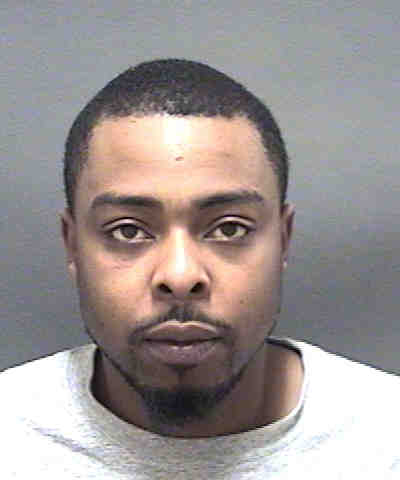} \\ \footnotesize Rank-one non-mate\\ from gallery of 500 \\ 286636\_02M25.JPG}} &
\fbox{\parbox{0.17\textwidth}{\centering \includegraphics[width=\linewidth]{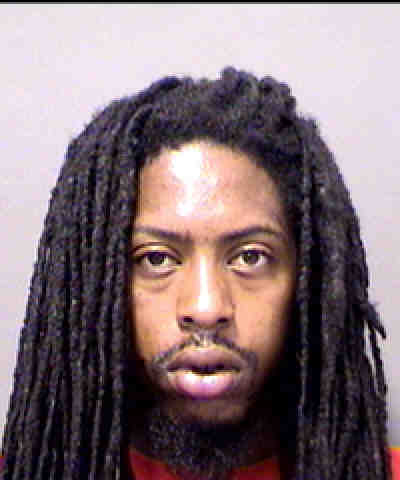} \\ \footnotesize Rank-one non-mate\\ from gallery of 5,000 \\ 237800\_17M29.JPG}} \\

\fbox{\parbox{0.17\textwidth}{\centering \includegraphics[width=\linewidth]{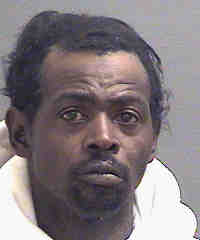} \\ \footnotesize Filler image \#2 \\ 090888\_04M41.JPG}} &
\fbox{\parbox{0.17\textwidth}{\centering \includegraphics[width=\linewidth]{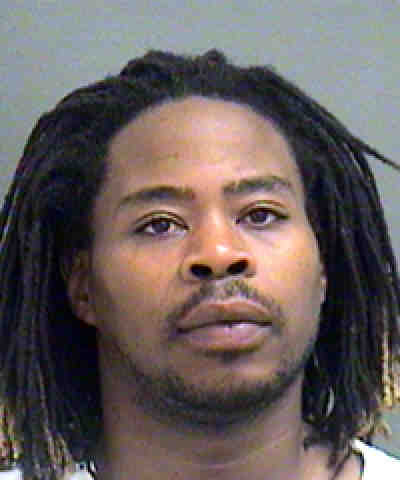} \\ \footnotesize Filler image \#2 \\ 152953\_06M36.JPG}} &
\fbox{\parbox{0.17\textwidth}{\centering \includegraphics[width=\linewidth]{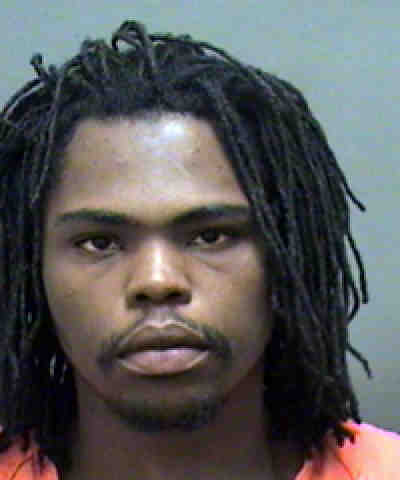} \\ \footnotesize Filler image \#3 \\ 326599\_01M20.JPG}} &
\fbox{\parbox{0.17\textwidth}{\centering \includegraphics[width=\linewidth]{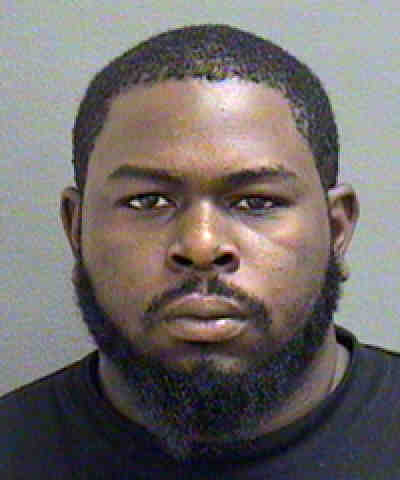} \\ \footnotesize Filler image \#4 \\ 343629\_01M25.JPG}} &
\fbox{\parbox{0.17\textwidth}{\centering \includegraphics[width=\linewidth]{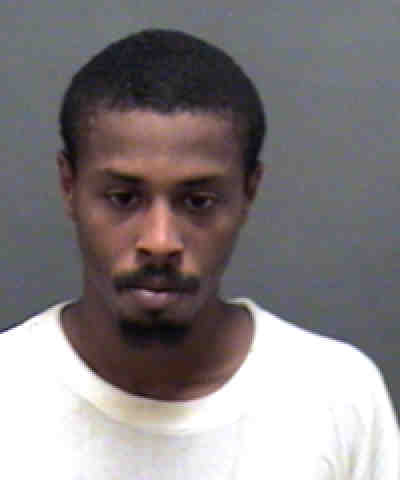} \\ \footnotesize filler image \#5 \\ 395228\_00M27.JPG}} \\

\fbox{\parbox{0.17\textwidth}{\centering \includegraphics[width=\linewidth]{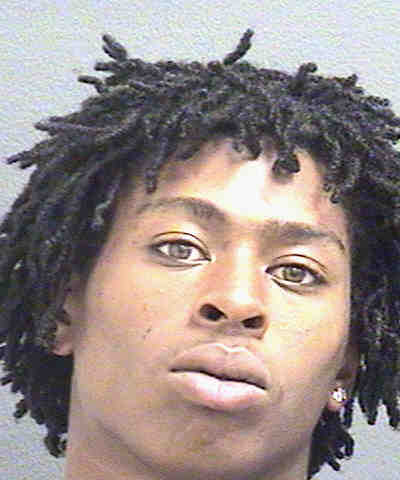} \\ \footnotesize Memory acquisition task\\ mated image \#1 \\ 259095\_01M21.JPG}} &
\fbox{\parbox{0.17\textwidth}{\centering \includegraphics[width=\linewidth]{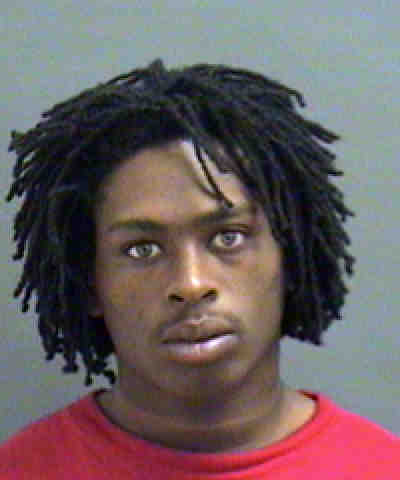} \\ \footnotesize Memory acquisition task\\ mated image \#2 \\ 259095\_02M22.JPG}} &
\fbox{\parbox{0.17\textwidth}{\centering \includegraphics[width=\linewidth]{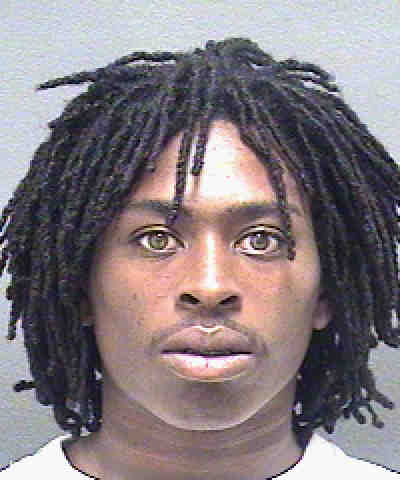} \\ \footnotesize Memory acquisition task\\ mated image \#3 \\ 259095\_03M23.JPG}} &
\fbox{\parbox{0.17\textwidth}{\centering \includegraphics[width=\linewidth]{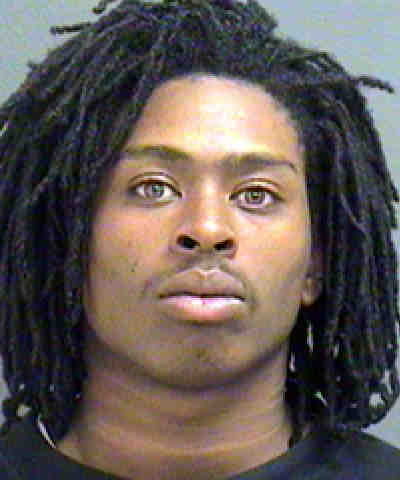} \\ \footnotesize Memory acquisition task\\ mated image \#4 \\ 259095\_05M23.JPG}} &
\\

\fbox{\parbox{0.17\textwidth}{\centering \includegraphics[width=\linewidth]{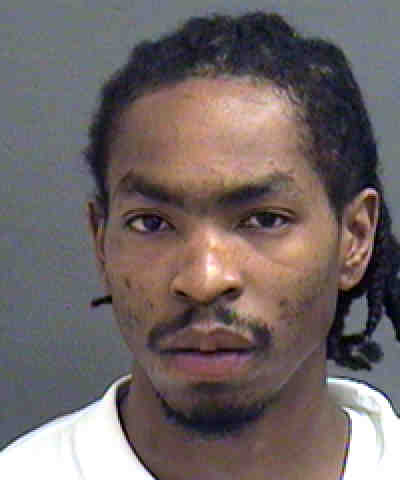} \\ \footnotesize Memory acquisition task\\ non-mated image \#1 \\ 236691\_08M27.JPG}} &
\fbox{\parbox{0.17\textwidth}{\centering \includegraphics[width=\linewidth]{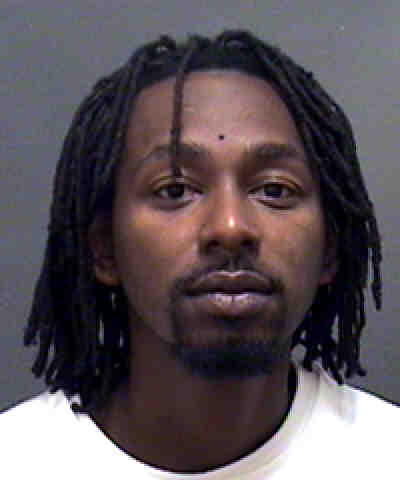} \\ \footnotesize memory acquisition task\\ non-mated image \#2 \\ 241997\_06M27.JPG}} &
\fbox{\parbox{0.17\textwidth}{\centering \includegraphics[width=\linewidth]{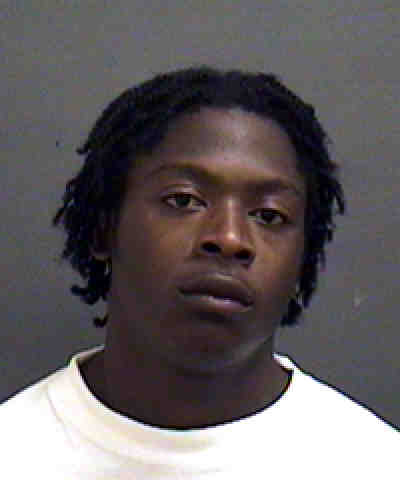} \\ \footnotesize memory acquisition task\\ non-mated image \#3 \\ 311508\_01M18.JPG}} &
\fbox{\parbox{0.17\textwidth}{\centering \includegraphics[width=\linewidth]{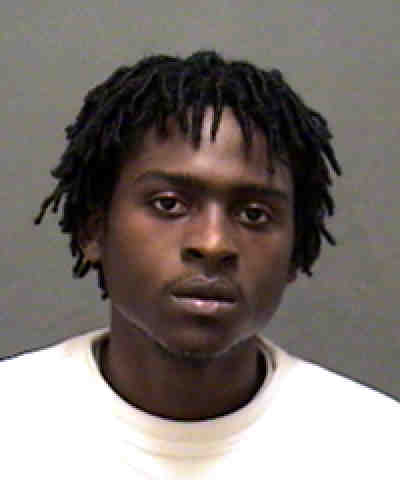} \\ \footnotesize memory acquisition task\\ non-mated image \#4 \\ 357328\_06M22.JPG}} &
\\
\end{tabular}

\caption{“Suspect 4” images used with six-pack photo lineups in experiment. Top row: image of suspect 1 used as probe to search against galleries, mated image for target present gallery, and rank-one non-mated images from galleries of varying sizes. Second row: filler images. Rows 3 and 4: images used in face memory acquisition task.}

\label{fig:suspect4} 
\end{figure*}

\begin{figure*}[t] 
\centering
\renewcommand{\arraystretch}{1.8} 
\setlength{\tabcolsep}{2pt}       

\begin{tabular}{ccccc}
\fbox{\parbox{0.17\textwidth}{\centering \includegraphics[width=\linewidth]{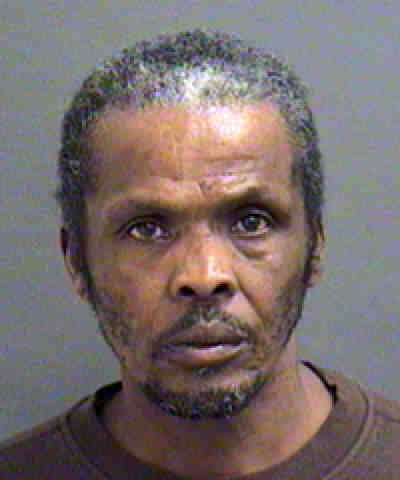} \\ \footnotesize Probe image \\ 051169\_05M54.JPG}} &
\fbox{\parbox{0.17\textwidth}{\centering \includegraphics[width=\linewidth]{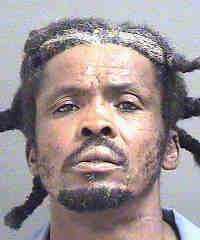} \\ \footnotesize Mated\\ gallery image \\ 051169\_03M49.JPG}} &
\fbox{\parbox{0.17\textwidth}{\centering \includegraphics[width=\linewidth]{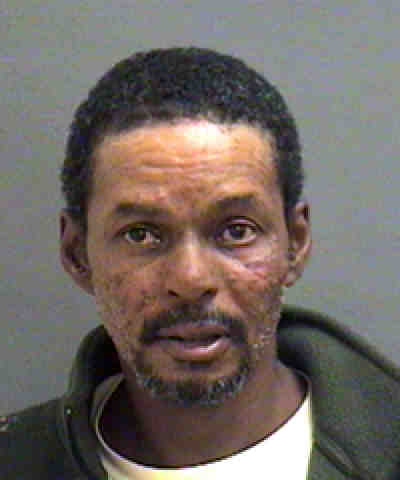} \\ \footnotesize Rank-one non-mate\\ from gallery of 500 \\ 075582\_00M43.JPG}} &
\fbox{\parbox{0.17\textwidth}{\centering \includegraphics[width=\linewidth]{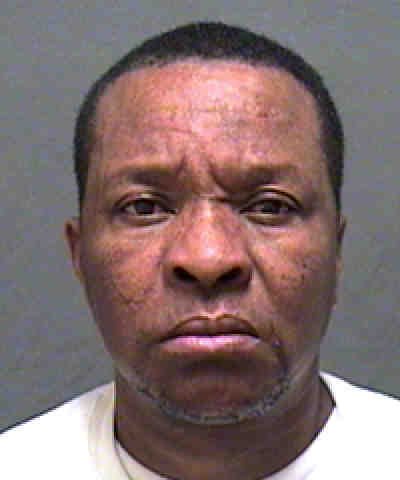} \\ \footnotesize Rank-one non-mate\\ from gallery of 5,000 \\ 060178\_12M49.JPG}} &
\fbox{\parbox{0.17\textwidth}{\centering \includegraphics[width=\linewidth]{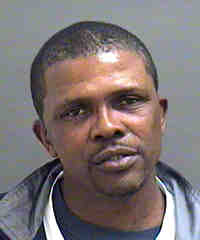} \\ \footnotesize Rank-one non-mate\\ from gallery of 24,000 \\ 11053626\_03M44}} \\

\fbox{\parbox{0.17\textwidth}{\centering \includegraphics[width=\linewidth]{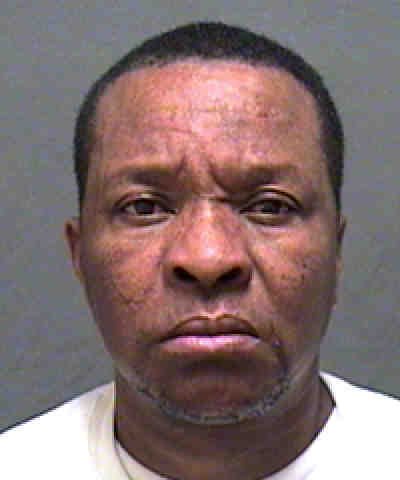} \\ \footnotesize Filler image \#1 \\ 060178\_12M49.JPG}} &
\fbox{\parbox{0.17\textwidth}{\centering \includegraphics[width=\linewidth]{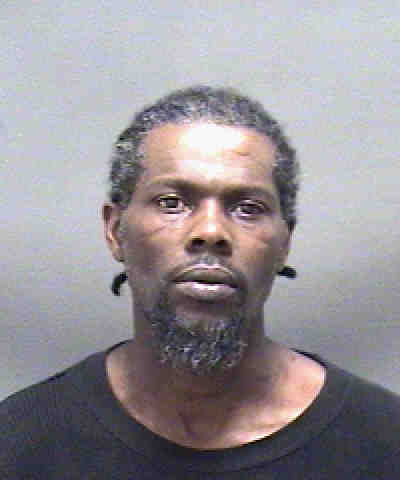} \\ \footnotesize Filler image \#2 \\ 098689\_06M39.JPG}} &
\fbox{\parbox{0.17\textwidth}{\centering \includegraphics[width=\linewidth]{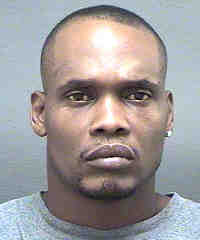} \\ \footnotesize Filler image \#3 \\ 102843\_01M35.JPG}} &
\fbox{\parbox{0.17\textwidth}{\centering \includegraphics[width=\linewidth]{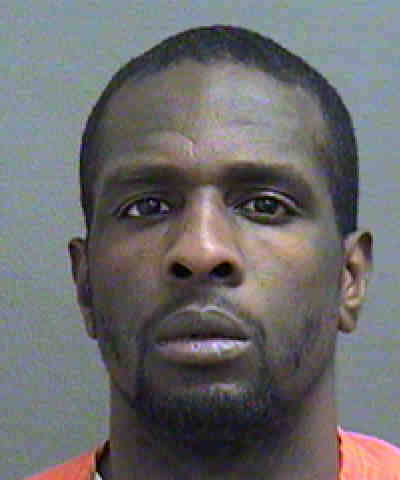} \\ \footnotesize Filler image \#4 \\ 242656\_09M26.JPG}} &
\fbox{\parbox{0.17\textwidth}{\centering \includegraphics[width=\linewidth]{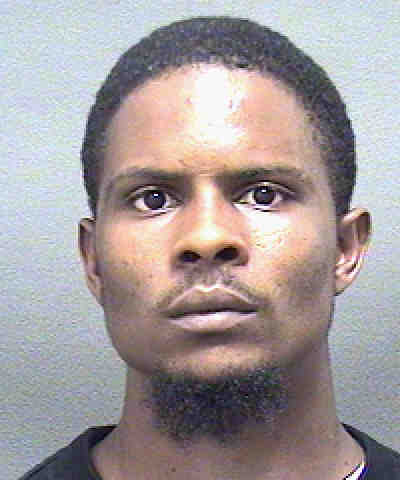} \\ \footnotesize Filler image \#5 \\ 286210\_03M22.JPG}} \\

\fbox{\parbox{0.17\textwidth}{\centering \includegraphics[width=\linewidth]{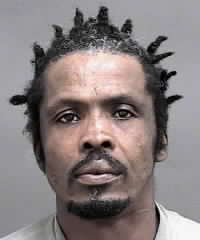} \\ \footnotesize Memory acquisition task\\ mated image \#1 \\ 051169\_00M46.JPG}} &
\fbox{\parbox{0.17\textwidth}{\centering \includegraphics[width=\linewidth]{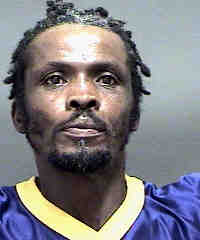} \\ \footnotesize Memory acquisition task\\ mated image \#2 \\ 051169\_01M47.JPG}} &
\fbox{\parbox{0.17\textwidth}{\centering \includegraphics[width=\linewidth]{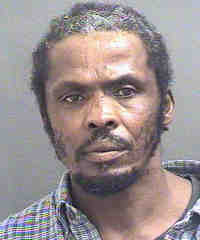} \\ \footnotesize Memory acquisition task\\ mated image \#3 \\ 051169\_02M48.JPG}} &
\fbox{\parbox{0.17\textwidth}{\centering \includegraphics[width=\linewidth]{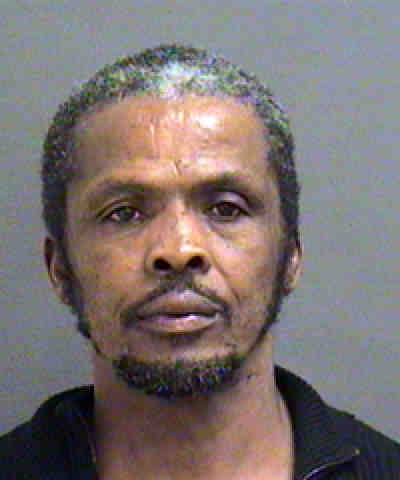} \\ \footnotesize Memory acquisition task\\ mated image \#4 \\ 051169\_04M53.JPG}} &
\\

\fbox{\parbox{0.17\textwidth}{\centering \includegraphics[width=\linewidth]{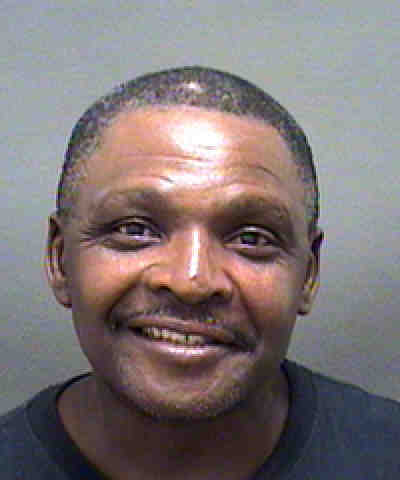} \\ \footnotesize Memory acquisition task\\ non-mated image \#1 \\ 043366\_01M53.JPG}} &
\fbox{\parbox{0.17\textwidth}{\centering \includegraphics[width=\linewidth]{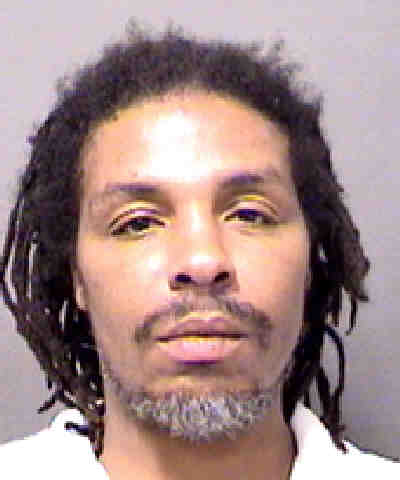} \\ \footnotesize Memory acquisition task\\ non-mated image \#2 \\ 096544\_09M43.JPG}} &
\fbox{\parbox{0.17\textwidth}{\centering \includegraphics[width=\linewidth]{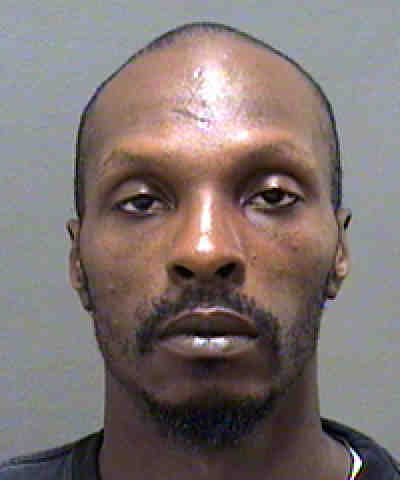} \\ \footnotesize Memory acquisition task\\ non-mated image \#3 \\ 132990\_05M35.JPG}} &
\fbox{\parbox{0.17\textwidth}{\centering \includegraphics[width=\linewidth]{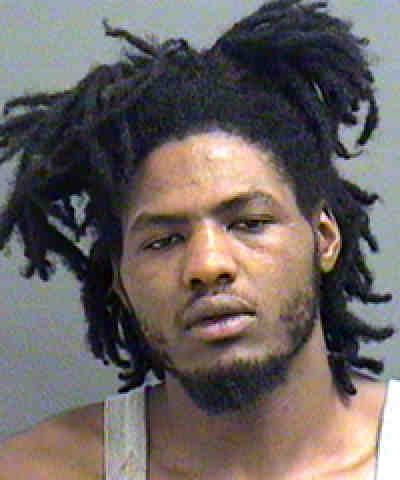} \\ \footnotesize Memory acquisition task\\ non-mated image \#4 \\ 252351\_02M23.JPG}} &
\\
\end{tabular}

\caption{“Suspect 5” images used with six-pack photo lineups in experiment. Top row: image of suspect 1 used as probe to search against galleries, mated image for target present gallery, and rank-one non-mated images from galleries of varying sizes. Second row: filler images. Rows 3 and 4: images used in face memory acquisition task.}

\label{fig:suspect5} 
\end{figure*}

\begin{figure*}[t] 
\centering
\renewcommand{\arraystretch}{1.8} 
\setlength{\tabcolsep}{2pt}       

\begin{tabular}{ccccc}
\fbox{\parbox{0.17\textwidth}{\centering \includegraphics[width=\linewidth]{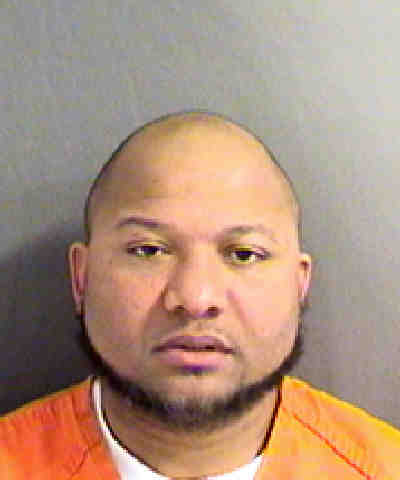} \\ \footnotesize Probe image \\ 190686\_07M35.JPG}} &
\fbox{\parbox{0.17\textwidth}{\centering \includegraphics[width=\linewidth]{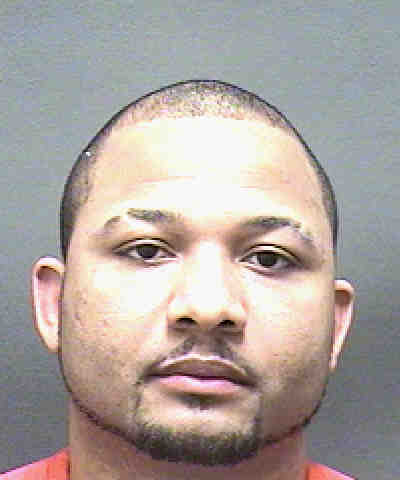} \\ \footnotesize Mated\\ gallery image \\ 190686\_02M29.JPG}} &
\fbox{\parbox{0.17\textwidth}{\centering \includegraphics[width=\linewidth]{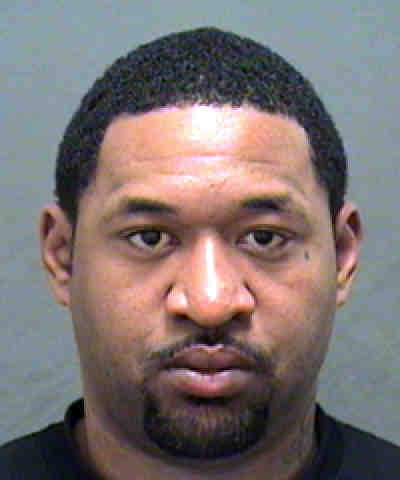} \\ \footnotesize Rank-one non-mate\\ from gallery of 500 \\ 305875\_04M28.JPG}} &
\fbox{\parbox{0.17\textwidth}{\centering \includegraphics[width=\linewidth]{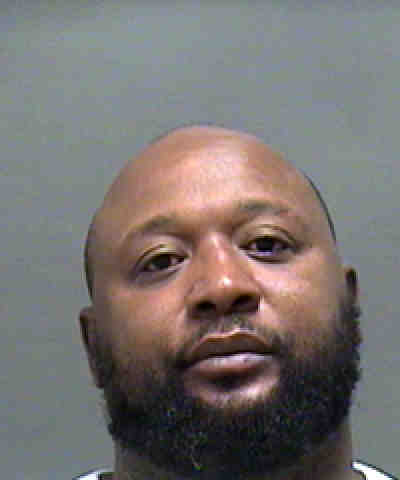} \\ \footnotesize Rank-one non-mate\\ from gallery of 5,000 \\ 120115\_07M38.JPG}} &
\fbox{\parbox{0.17\textwidth}{\centering \includegraphics[width=\linewidth]{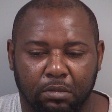} \\ \footnotesize Rank-one non-mate\\ from gallery of 24,000 \\ 225291530\_17M4}} \\

\fbox{\parbox{0.17\textwidth}{\centering \includegraphics[width=\linewidth]{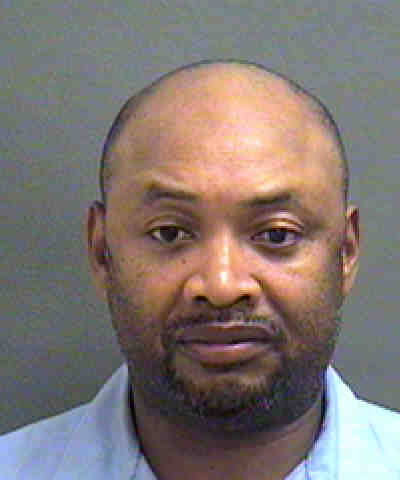} \\ \footnotesize Filler image \#1 \\ 061646\_03M50.JPG}} &
\fbox{\parbox{0.17\textwidth}{\centering \includegraphics[width=\linewidth]{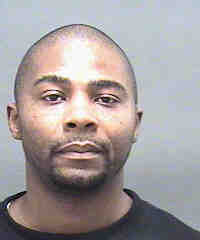} \\ \footnotesize Filler image \#2 \\ 099237\_02M34.JPG}} &
\fbox{\parbox{0.17\textwidth}{\centering \includegraphics[width=\linewidth]{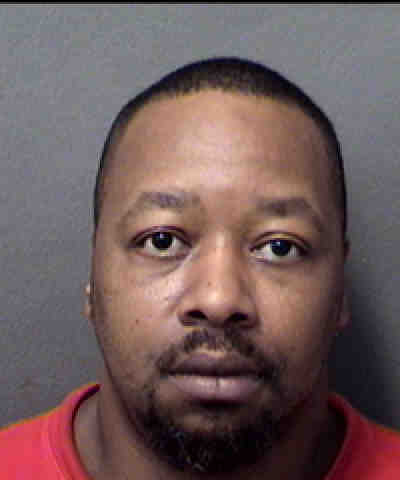} \\ \footnotesize Filler image \#3 \\ 153897\_04M41.JPG}} &
\fbox{\parbox{0.17\textwidth}{\centering \includegraphics[width=\linewidth]{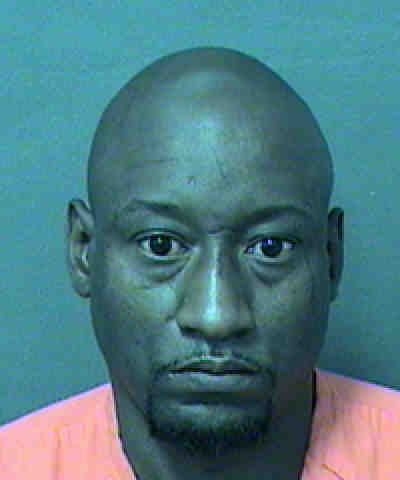} \\ \footnotesize Filler image \#4 \\ 178384\_11M37.JPG}} &
\fbox{\parbox{0.17\textwidth}{\centering \includegraphics[width=\linewidth]{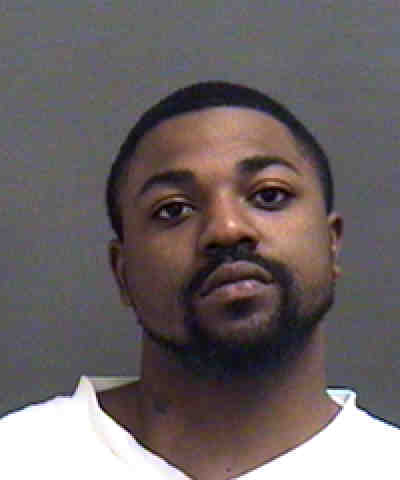} \\ \footnotesize Filler image \#5 \\ 325547\_04M24.JPG}} \\

\fbox{\parbox{0.17\textwidth}{\centering \includegraphics[width=\linewidth]{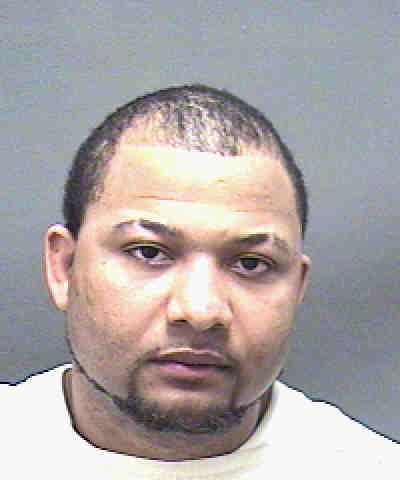} \\ \footnotesize Memory acquisition task\\ mated image \#1 \\ 190686\_01M29.JPG}} &
\fbox{\parbox{0.17\textwidth}{\centering \includegraphics[width=\linewidth]{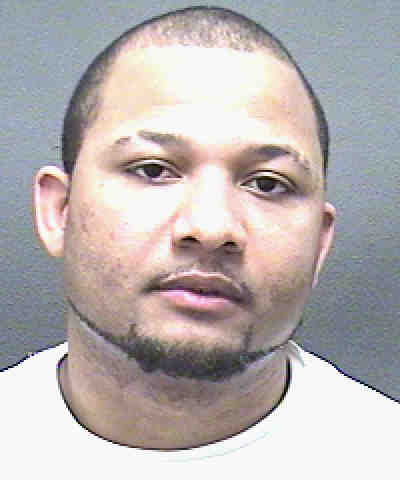} \\ \footnotesize Memory acquisition task\\ mated image \#2 \\ 190686\_03M29.JPG}} &
\fbox{\parbox{0.17\textwidth}{\centering \includegraphics[width=\linewidth]{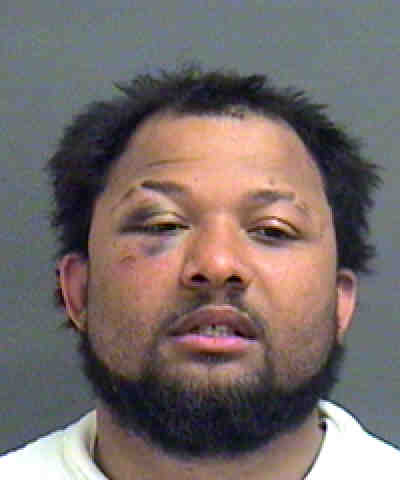} \\ \footnotesize Memory acquisition task\\ mated image \#3 \\ 190686\_05M30.JPG}} &
\fbox{\parbox{0.17\textwidth}{\centering \includegraphics[width=\linewidth]{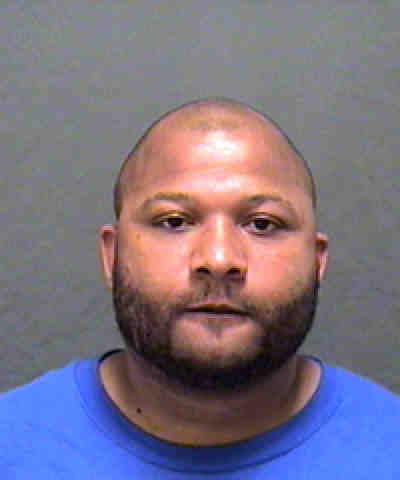} \\ \footnotesize Memory acquisition task\\ mated image \#4 \\ 190686\_06M34.JPG}} &
\\

\fbox{\parbox{0.17\textwidth}{\centering \includegraphics[width=\linewidth]{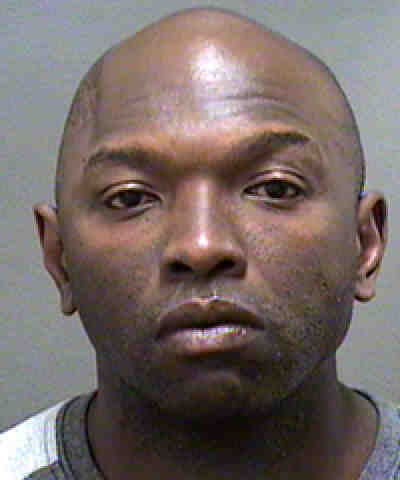} \\ \footnotesize Memory acquisition task\\ non-mated image \#1 \\ 095470\_21M41.JPG}} &
\fbox{\parbox{0.17\textwidth}{\centering \includegraphics[width=\linewidth]{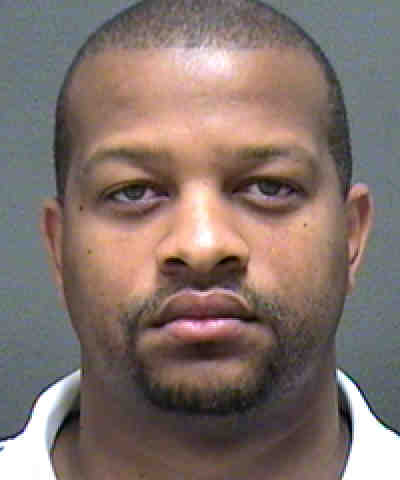} \\ \footnotesize Memory acquisition task\\ non-mated image \#2 \\ 168227\_00M33.JPG}} &
\fbox{\parbox{0.17\textwidth}{\centering \includegraphics[width=\linewidth]{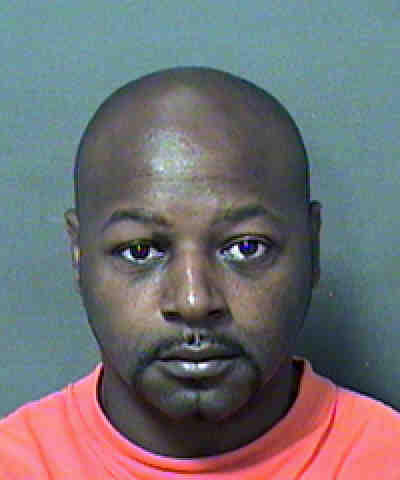} \\ \footnotesize Memory acquisition task\\ non-mated image \#3 \\ 313549\_14M34.JPG}} &
\fbox{\parbox{0.17\textwidth}{\centering \includegraphics[width=\linewidth]{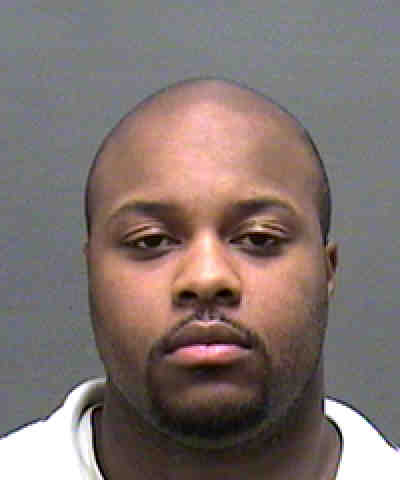} \\ \footnotesize Memory acquisition task\\ non-mated image \#4 \\ 366203\_00M23.JPG}} &
\\
\end{tabular}

\caption{“Suspect 6” images used with six-pack photo lineups in experiment. Top row: image of suspect 1 used as probe to search against galleries, mated image for target present gallery, and rank-one non-mated images from galleries of varying sizes. Second row: filler images. Rows 3 and 4: images used in face memory acquisition task.}

\label{fig:suspect6} 
\end{figure*}

\begin{figure*}[t] 
\centering
\renewcommand{\arraystretch}{1.8} 
\setlength{\tabcolsep}{2pt}       

\begin{tabular}{ccccc}
\fbox{\parbox{0.17\textwidth}{\centering \includegraphics[width=\linewidth]{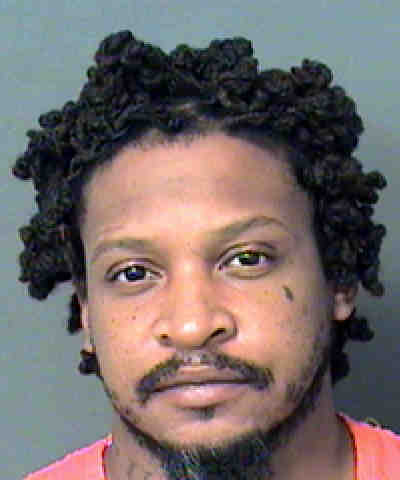} \\ \footnotesize Probe image \\ 201792\_12M33.JPG}} &
\fbox{\parbox{0.17\textwidth}{\centering \includegraphics[width=\linewidth]{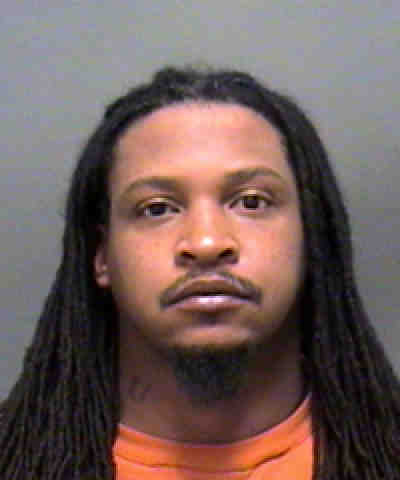} \\ \footnotesize Mated\\ gallery image \\ 201792\_07M30.JPG}} &
\fbox{\parbox{0.17\textwidth}{\centering \includegraphics[width=\linewidth]{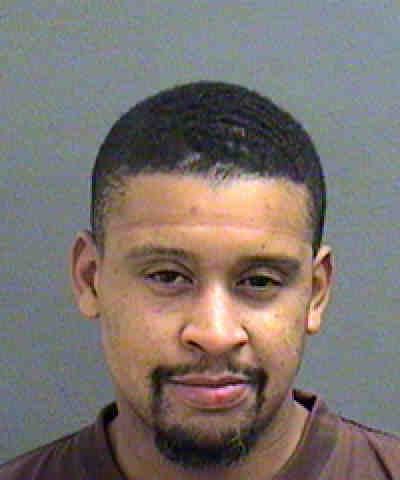} \\ \footnotesize Rank-one non-mate\\ from gallery of 500 \\ 344162\_01M37.JPG}} &
\fbox{\parbox{0.17\textwidth}{\centering \includegraphics[width=\linewidth]{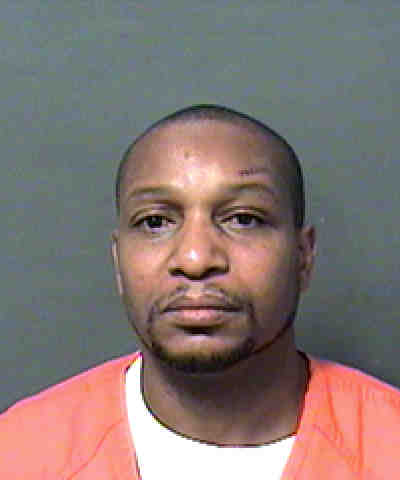} \\ \footnotesize Rank-one non-mate\\ from gallery of 5,000 \\ 080991\_06M45.JPG}} &
\fbox{\parbox{0.17\textwidth}{\centering \includegraphics[width=\linewidth]{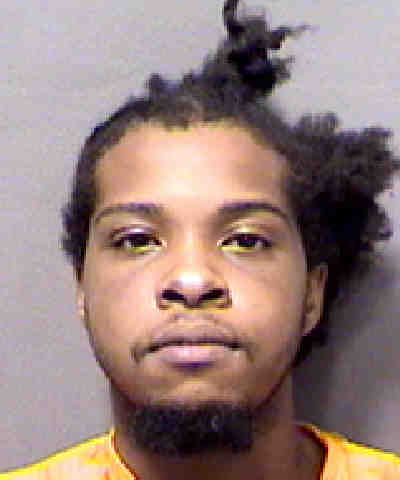} \\ \footnotesize Rank-one non-mate\\ from gallery of 24,000 \\ 1336732\_02M23}} \\

\fbox{\parbox{0.17\textwidth}{\centering \includegraphics[width=\linewidth]{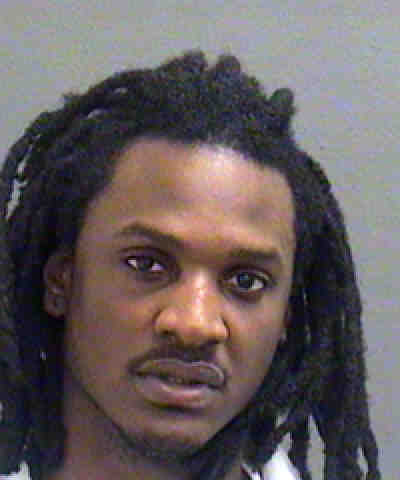} \\ \footnotesize Filler image \#1 \\ 249460\_10M26.JPG}} &
\fbox{\parbox{0.17\textwidth}{\centering \includegraphics[width=\linewidth]{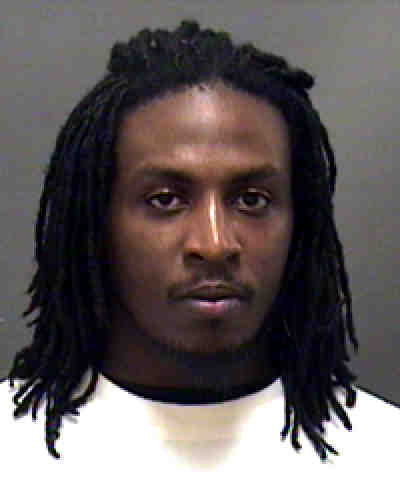} \\ \footnotesize Filler image \#2 \\ 280477\_14M23.JPG}} &
\fbox{\parbox{0.17\textwidth}{\centering \includegraphics[width=\linewidth]{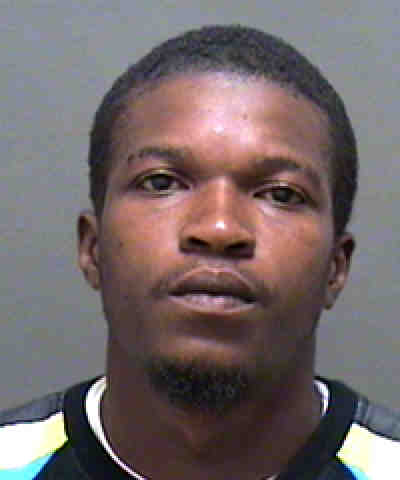} \\ \footnotesize Filler image \#3 \\ 290299\_11M24.JPG}} &
\fbox{\parbox{0.17\textwidth}{\centering \includegraphics[width=\linewidth]{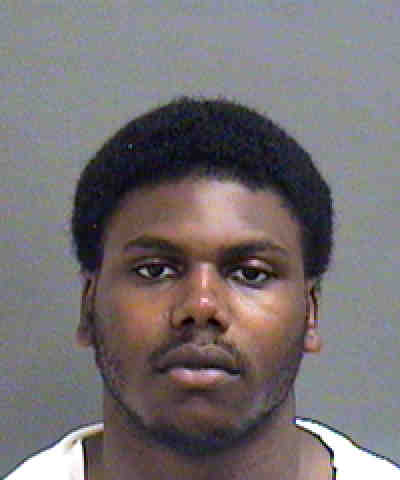} \\ \footnotesize Filler image \#4 \\ 305682\_02M20.JPG}} &
\fbox{\parbox{0.17\textwidth}{\centering \includegraphics[width=\linewidth]{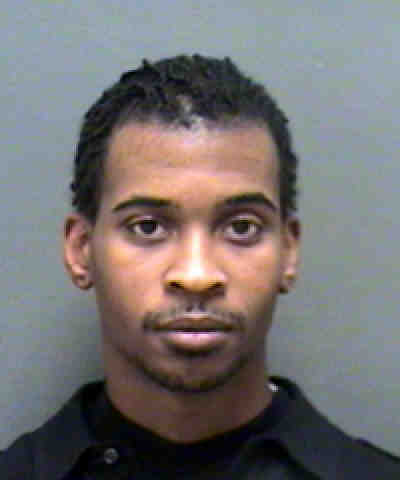} \\ \footnotesize Filler image \#5 \\ 389824\_01M23.JPG}} \\

\fbox{\parbox{0.17\textwidth}{\centering \includegraphics[width=\linewidth]{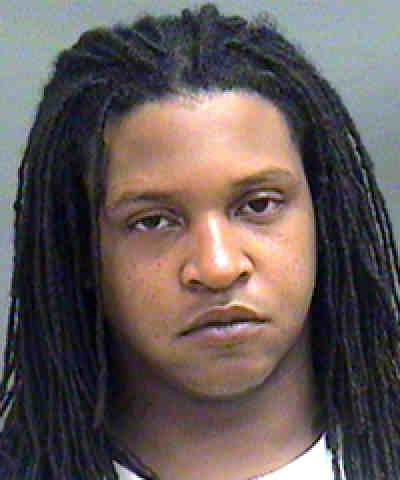} \\ \footnotesize Memory acquisition task\\ mated image \#1 // 201792\_00M29.JPG}} &
\fbox{\parbox{0.17\textwidth}{\centering \includegraphics[width=\linewidth]{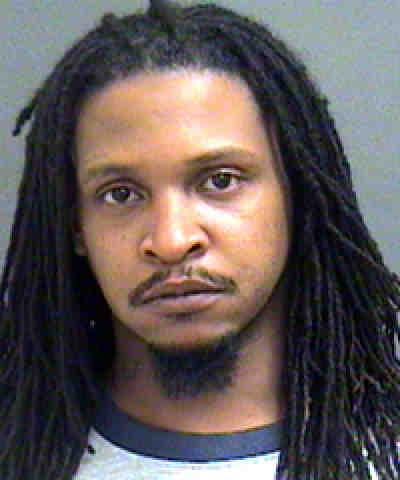} \\ \footnotesize Memory acquisition task\\ mated image \#2 // 201792\_03M29.JPG}} &
\fbox{\parbox{0.17\textwidth}{\centering \includegraphics[width=\linewidth]{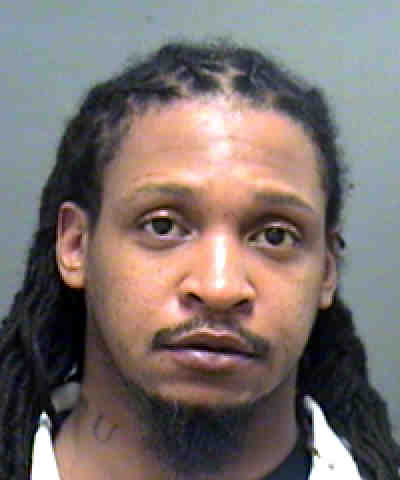} \\ \footnotesize Memory acquisition task\\ mated image \#3 \\ 201792\_10M31.JPG}} &
\fbox{\parbox{0.17\textwidth}{\centering \includegraphics[width=\linewidth]{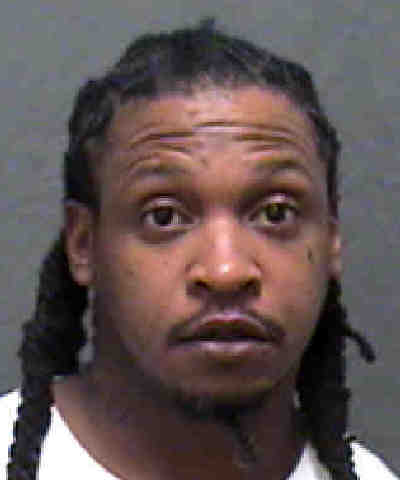} \\ \footnotesize Memory acquisition task\\ mated image \#4 \\ 201792\_11M33.JPG}} &
\\

\fbox{\parbox{0.17\textwidth}{\centering \includegraphics[width=\linewidth]{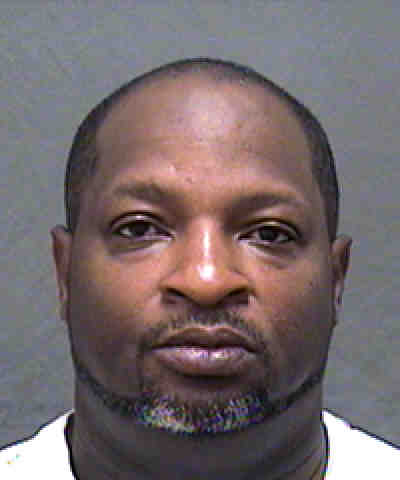} \\ \footnotesize Memory acquisition task\\ non-mated image \#1 \\ 077903\_05M45.JPG }} &
\fbox{\parbox{0.17\textwidth}{\centering \includegraphics[width=\linewidth]{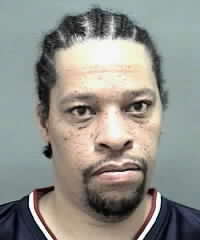} \\ \footnotesize Memory acquisition task\\ non-mated image \#2 \\ 098190\_00M42.JPG}} &
\fbox{\parbox{0.17\textwidth}{\centering \includegraphics[width=\linewidth]{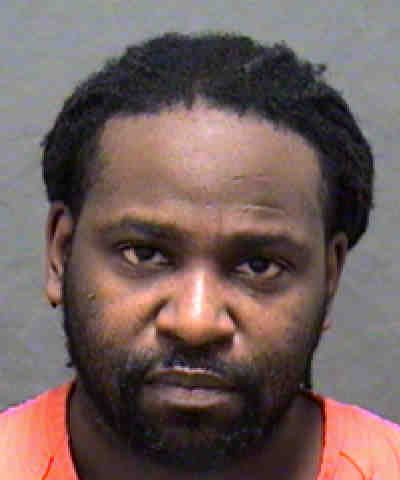} \\ \footnotesize Memory acquisition task\\ non-mated image \#3 \\ 235001\_12M31.JPG}} &
\fbox{\parbox{0.17\textwidth}{\centering \includegraphics[width=\linewidth]{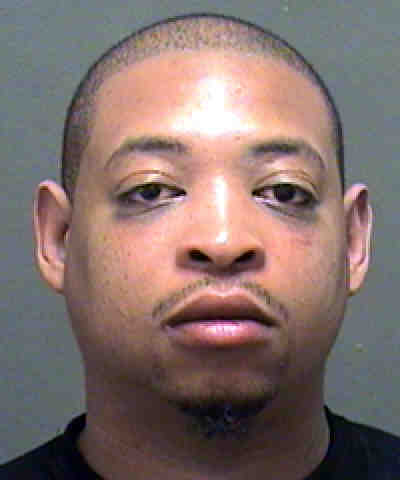} \\ \footnotesize Memory acquisition task\\ non-mated image \#4 \\ 265915\_09M27.JPG}} &
\\
\end{tabular}

\caption{“Suspect 7” images used with six-pack photo lineups in experiment. Top row: image of suspect 1 used as probe to search against galleries, mated image for target present gallery, and rank-one non-mated images from galleries of varying sizes. Second row: filler images. Rows 3 and 4: images used in face memory acquisition task.}

\label{fig:suspect7} 
\end{figure*}

\begin{figure*}[t] 
\centering
\renewcommand{\arraystretch}{1.8} 
\setlength{\tabcolsep}{2pt}       

\begin{tabular}{ccccc}
\fbox{\parbox{0.17\textwidth}{\centering \includegraphics[width=\linewidth]{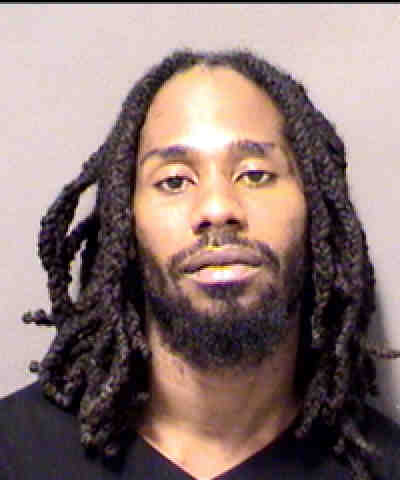} \\ \footnotesize Probe image \\ 296102\_11M27.JPG}} &
\fbox{\parbox{0.17\textwidth}{\centering \includegraphics[width=\linewidth]{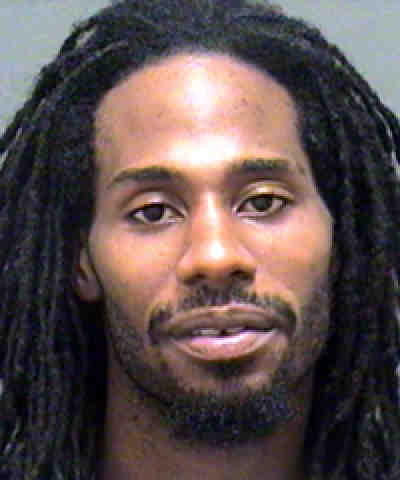} \\ \footnotesize Mated\\ gallery image \\ 296102\_08M26.JPG}} &
\fbox{\parbox{0.17\textwidth}{\centering \includegraphics[width=\linewidth]{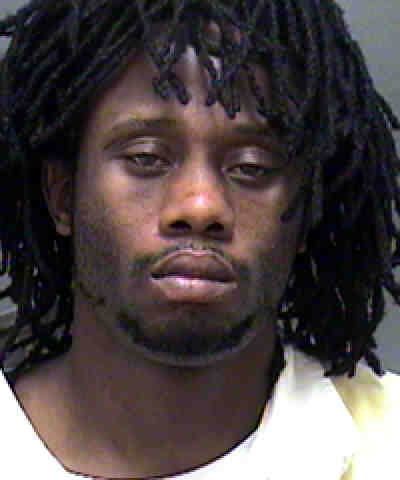} \\ \footnotesize Rank-one non-mate\\ from gallery of 500 \\ 205646\_07M29.JPG}} &
\fbox{\parbox{0.17\textwidth}{\centering \includegraphics[width=\linewidth]{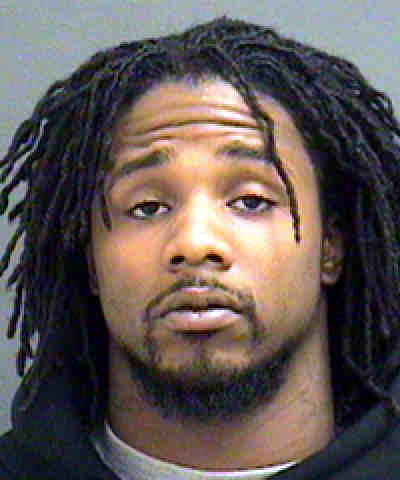} \\ \footnotesize Rank-one non-mate\\ from gallery of 5,000 \\ 239733\_10M24.JPG}} &
\fbox{\parbox{0.17\textwidth}{\centering \includegraphics[width=\linewidth]{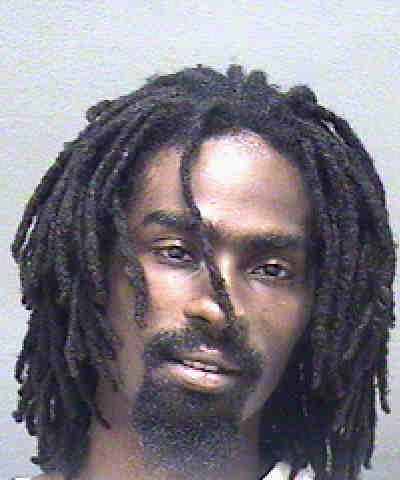} \\ \footnotesize Rank-one non-mate\\ from gallery of 24,000 \\ 11195273\_00M29}} \\

\fbox{\parbox{0.17\textwidth}{\centering \includegraphics[width=\linewidth]{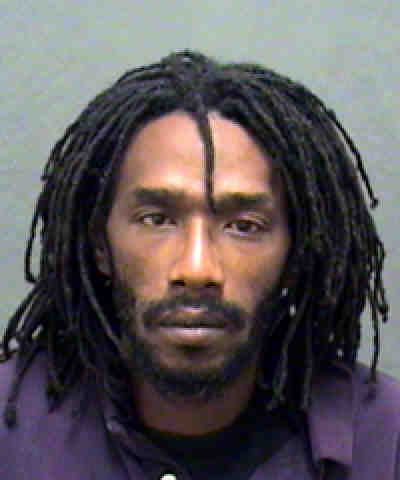} \\ \footnotesize Filler image \#1 \\ 120082\_21M39.JPG}} &
\fbox{\parbox{0.17\textwidth}{\centering \includegraphics[width=\linewidth]{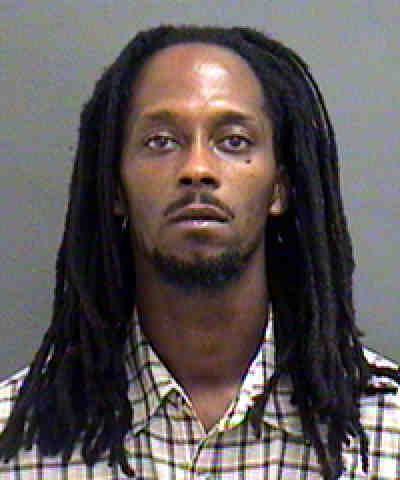} \\ \footnotesize Filler image \#2 \\ 174774\_01M31.JPG}} &
\fbox{\parbox{0.17\textwidth}{\centering \includegraphics[width=\linewidth]{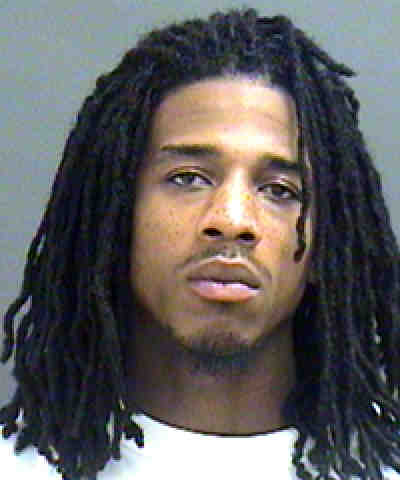} \\ \footnotesize Filler image \#3 \\ 230999\_08M25.JPG}} &
\fbox{\parbox{0.17\textwidth}{\centering \includegraphics[width=\linewidth]{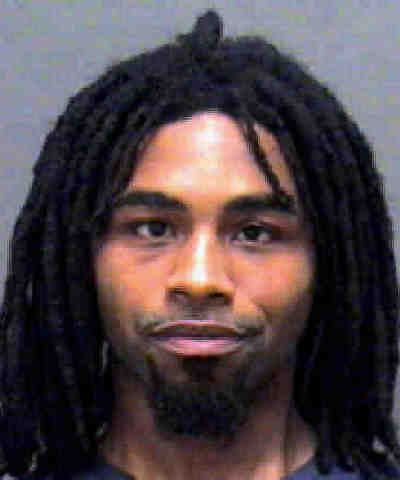} \\ \footnotesize Filler image \#4 \\ 237577\_00M26.JPG}} &
\fbox{\parbox{0.17\textwidth}{\centering \includegraphics[width=\linewidth]{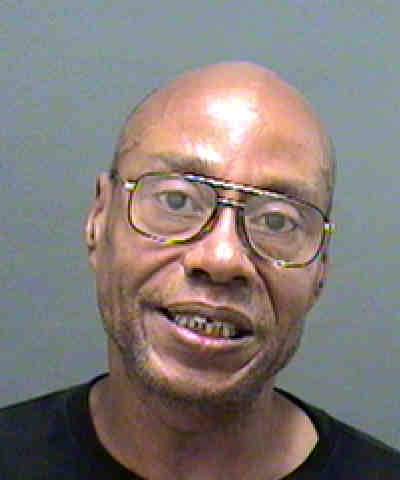} \\ \footnotesize Filler image \#5 \\ 350093\_03M53.JPG}} \\

\fbox{\parbox{0.17\textwidth}{\centering \includegraphics[width=\linewidth]{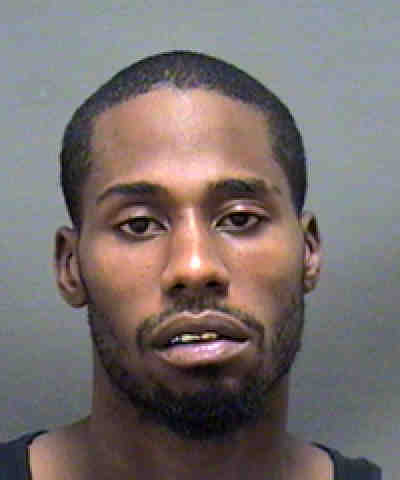} \\ \footnotesize Memory acquisition task\\ mated image \#1 \\ 296102\_04M22.JPG}} &
\fbox{\parbox{0.17\textwidth}{\centering \includegraphics[width=\linewidth]{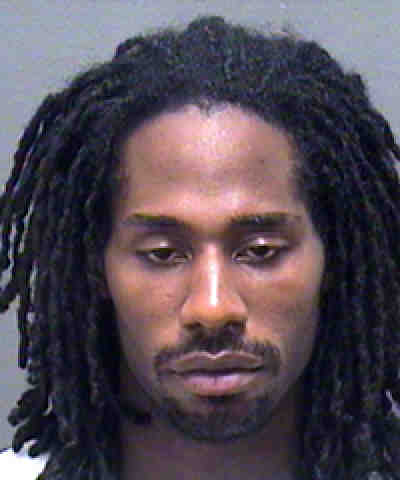} \\ \footnotesize Memory acquisition task\\ mated image \#2 \\ 296102\_07M25.JPG}} &
\fbox{\parbox{0.17\textwidth}{\centering \includegraphics[width=\linewidth]{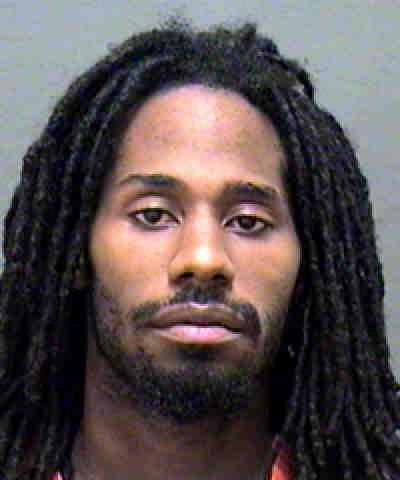} \\ \footnotesize Memory acquisition task\\ mated image \#3 \\ 296102\_09M26.JPG}} &
\fbox{\parbox{0.17\textwidth}{\centering \includegraphics[width=\linewidth]{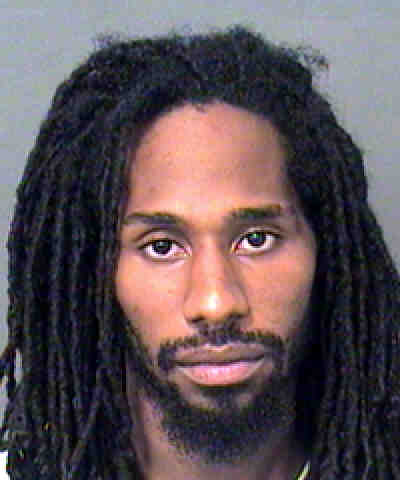} \\ \footnotesize Memory acquisition task\\ mated image \#4 \\ 296102\_10M26.JPG}} &
\\

\fbox{\parbox{0.17\textwidth}{\centering \includegraphics[width=\linewidth]{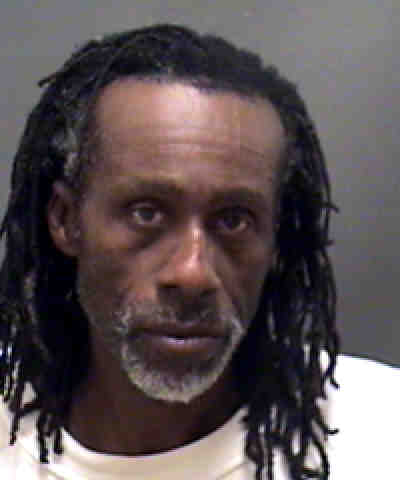} \\ \footnotesize Memory acquisition task\\ non-mated image \#1 \\ 147613\_06M48.JPG}} &
\fbox{\parbox{0.17\textwidth}{\centering \includegraphics[width=\linewidth]{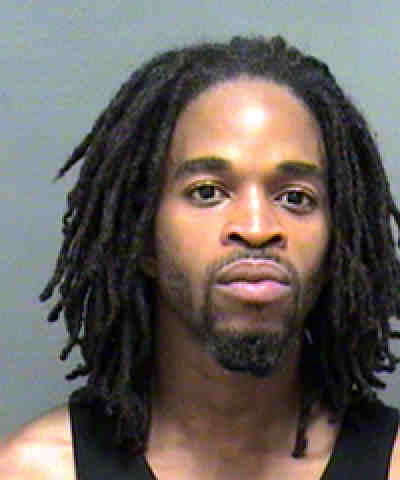} \\ \footnotesize memory acquisition task\\ non-mated image \#2 \\ 187674\_08M32.JPG}} &
\fbox{\parbox{0.17\textwidth}{\centering \includegraphics[width=\linewidth]{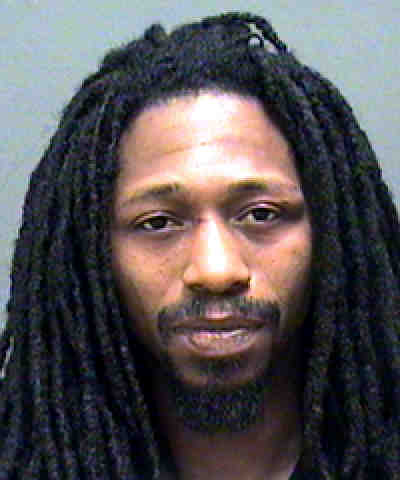} \\ \footnotesize memory acquisition task\\ non-mated image \#3 \\ 213195\_08M31.JPG}} &
\fbox{\parbox{0.17\textwidth}{\centering \includegraphics[width=\linewidth]{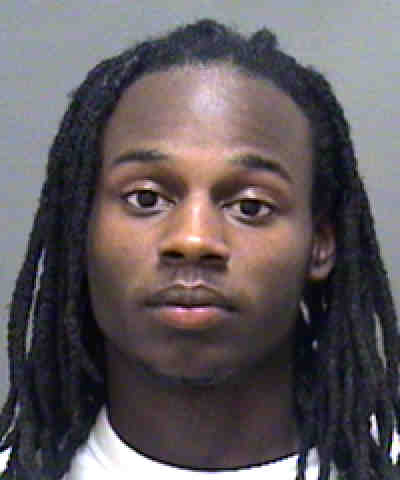} \\ \footnotesize memory acquisition task\\ non-mated image \#4 \\ 334780\_06M18.JPG}} &
\\
\end{tabular}

\caption{“Suspect 8” images used with six-pack photo lineups in experiment. Top row: image of suspect 1 used as probe to search against galleries, mated image for target present gallery, and rank-one non-mated images from galleries of varying sizes. Second row: filler images. Rows 3 and 4: images used in face memory acquisition task.}

\label{fig:suspect8} 
\end{figure*}

\clearpage

\end{document}